# Automated Detection and Structuring of Social Tipping Point Evidence in Climate related Documents: A Modular AI Framework

Kavindu Perera[1], Mohammad Abaeiani[1], Ekaterina Gilman[1], Lauri Loven[1], Mourad Oussalah[2], Tassos Kanellos[3], Beatrice Gobbo[7], Dante Adami[4], Nicolo` Ferriani[4], Maximiliano Romero[7], Pierre Rossel[8], Marc Bonazountas[6], Christina Deligianni[5], Nikos Xyderis[6], Artur Bogucki[9], Lampros Argyriou[3], Prasasthy Balasubramanian[1*]

[1]Center for Applied Computing, University of Oulu, Oulu, 90570, North Ostrobothnia, Finland.
[2]Center for Machine Vision and Signal Analysis, University of Oulu, Oulu, 90570, North Ostrobothnia, Finland.
[3]ITML CY, 185 Arch. Makariou III Ave, Limassol, 3095, Cyprus. [4]PredictBy Research and Consulting SL, Carrer d'Arag´o 383, Barcelona, 08013, Catalonia, Spain.
[5]Verimpact, Olympionikon 18A, Pikermi, 19009, Greece.
[6]Epsilon International Ltd, Tower Business Center (212), Tower Street, Swatar-Birkirkara, BKR 4013, Malta.
[7]Department of Design, Politecnico di Milano, Via Durando 10, Milan, 20158, Italy.
[8]Inspiring Futures Europe, Calle Espronceda, Madrid, 28003, Spain. [9]Centre for European Policy Studies (CEPS), Place du Congr`es 1, Brussels, 1000, Belgium.

*Corresponding author(s). E-mail(s): prasasthy.baalsubramanian@oulu.fi;

**Abstract**

The climate literature has grown faster than review teams can read it. That gap matters most for a concept like the environmental social tipping point, the threshold at which a small change triggers rapid, self-reinforcing change in a social system. Evidence of this kind of shift is usually contained in one or two paragraphs within a longer document. As a result, existing text mining toolswhich categorize entire documents by topic or highlight isolated claims-leave an expanding set of important evidence without any systematic method for discovery or organization. This paper presents an open and modular transformerbased framework that detects and structures social tipping point evidence at the passage level. The framework joins five components into a single deployable workflow: a DistilBERT boundary splitter for segmentation, an iteratively augmented RoBERTa classifier for detection, a Mistral 7B model that rewrites each detected passage for clarity, a LLaMA 3.2 3B model that rates the passage against five published social tipping point criteria, and a Milvus vector store for semantic retrieval. The system is wrapped in a Streamlit interface backed by MinIO object storage. Evaluated on a 163-passage benchmark labelled by GPT-4.1 and a 51-passage set reviewed by experts, the splitter surpassed three competing methods on a nine-metric composite score (6.137). The tuned RoBERTa model achieved 71.4 percent accuracy with a Cohen's kappa of 0.337 on the full benchmark, and 87.5 percent accuracy with a kappa of 0.742 on passages with labels, outperforming both a climate-focused model and untuned language models.

# 1 Introduction

Climate-related publications grew by roughly a factor of ten from 1991 to 2010, and this rate is still continuously increasing [1, 2]. As a result, manual review of existing literature cannot keep pace, and several surveys already identified this accumulation as a key barrier to scaling climate-evidence synthesis [3, 4]. Especially, from climate policy perspective, it is crucial to understand how small and possibly incremental change, often referred to as *social tipping threshold*, affects overall practices. More specifically, a *social tipping point* is a threshold at which a small change in a social system sets off a rapid, self-sustaining transition to a different state [5, 6]. Physical tipping elements such as ice-sheet collapse and Amazon dieback are critical Earth subsystems that maintain global stability so that an incremental increase of temperature, for instance, beyond a certain threshold can create devastating consequences to biodiversity ecosystem changes [7]. In general, social tipping describes an outcome that societies might want, and, in principle, it can be triggered deliberately through targeted action [6, 8]. Field experiments suggest that a committed minority of roughly a quarter of a group is sufficient to overturn an established norm [9], and recent work underscores a steep increase in research on social tipping through 2024 [10]. The economic consequences are well-documented [11], and the distinction between a tipping point that can be acted on and one that is merely crossed has only recently been made operational [12, 13]. Besides, the question whether such tipping points can be identified through automated scrutinization of climate literature is legitimate and worth exploring. Surveying existing climate text tools reveals two distinguished approaches, reflecting the top-down and bottom-up paths. The former constructs a collection consisting of a compilation of all individual documents, which is then followed by a topic-like analysis [2, 4, 14]. The second attempts to flag individual sentence (s) as a claim or a disclosure [15–17]. However, both methods face intrinsic constraints because impactful climate statements often cannot be captured as discussion topics and also fail to align with predefined ontology patterns for simple query matching. For example, the recent ClimateEval benchmark demonstrates that none of the existing climate-related text tasks operate at the passage level [18]. Similarly, the latest survey of language models on climate change mining highlights concept detection as a key open problem for climate mining [3]. Social-science research acknowledged that social tipping research remains in its infancy stage [19]. This is in part due to the domain expertise required to test the automatically generated hypotheses.

This paper presents an approach for building an open, modular system that finds and structures social tipping point evidence, *passage* by *passage*, across documents where climate knowledge is recorded, which include research papers, policy reports, online audiovisual content, and news. The system is built so that each component can be replaced independently as better tools become available, and so that the whole workflow can be deployed locally and self-managed, removing the dependency on closed commercial pipelines and the risks of vendor lock-in and data sovereignty loss that come with it. A research team can therefore use such a system to dynamically build and update a map of where the literature already documents the build-up toward a social tipping point, at the pace the literature is growing rather than years after the fact, and to export the resulting structured evidence in formats fit for downstream policy and research tasks.

To systematically evaluate the proposed framework and its individual components, this study addresses the following research questions:

- **RQ1**: Can passage-level document segmentation improve the identification of social tipping point (STP) evidence in climate-related documents?
- **RQ2**: Can transformer-based models enhanced through iterative augmentation achieve reliable STP classification under limited expert-labeled data conditions?
- **RQ3**: Can a modular AI framework support scalable detection, enhancement, and analysis of STP evidence for downstream climate intelligence applications?

In overall, the paper makes the following five contributions:

1. A passage-level framework for social tipping point detection that joins segmentation, classification, rewriting, and criterion-based rating into one open workflow is proposed.
2. A comparative evaluation of four climate-document segmentation strategies demonstrates that DistilBERT-based boundary detection provides the most effective balance between semantic coherence, coverage, and chunk quality.
3. An iteratively augmented RoBERTa-based classification approach designed for lowresource STP detection is put forward, together with a documented analysis of how targeted augmentation influences class balance and classification performance.
4. A structured qualifying factor generator, that rates detected passages against five criteria from the social tipping point literature, provides confidence-rated evidence summaries grounded in specific passage content.

5. A complete deployable system that wraps the four model components in a configurable Streamlit interface backed by MinIO object storage is devised and demonstrated, including an ONNX-compiled classifier and Milvus vector store, while the structured outputs are exported into JSON, CSV, and Excel formats for downstream research and policy work.

The remainder of the paper is organized as follows. Section 2 reviews existing work on social tipping points and on machine learning for climate evidence, and sets out the research gap. Section 3 describes the proposed framework, including the data, the model components, the system architecture, and the evaluation methodology. Section 4 reports the results. Section 5 discusses what the findings mean for the design of applied text mining systems and concludes.

# 2 Related work

## 2.1 Social tipping points and why they are hard to detect

Tipping-point theory originated in complex-systems research, where a tipping point is a threshold beyond which a system shifts rapidly and often irreversibly to a different state [7]. Lamberson and Page [20] gave an early formal account of the social version as a choice between two equilibria in a mixed population. Milkoreit et al. [5] reviewed five decades of tipping-point work across fields and reduced twenty-three definitional features to one working definition built on nonlinear change, positive feedback, and limited reversibility. Otto et al. [6] named six social tipping elements whose inner dynamics can drive rapid decarbonisation once the right intervention points are switched on, and T`abara et al. [21] described similar positive tipping across energy, cities, and political norms. Lenton et al. [12] set out tests that separate a tipping point people can act on from a threshold crossing that simply happens, and Winkelmann et al. [22] set out tests that separate social tipping from physical tipping. Field evidence has come from Centola et al. [9] on the twenty-five per cent threshold for norm change, Castilla-Rho et al. [23] on groundwater rules across thirty countries, Martin et al. [24] on transient social dynamics, and Peng and Bai [25] on peer-effect thresholds drawn from survey data.

Several recent reviews show where the research stands. Milkoreit [19] found that most social tipping work stays at the level of ideas because turning it into measured evidence is costly. Riekhof et al. [26] flagged a deeper problem, that whether a tipping point shows up at all depends on the time window used to look for it. Juhola et al. [27] placed social tipping inside a frame of systemic risk, and Froese et al. [28] offered a frame for the way social and ecological systems interact to produce tipping. Hodbod et al. [13] drew up rules for sound tipping-point case studies and stressed that a claim needs evidence of the feedback mechanism, not just evidence of fast change. Graham et al. [29] added intent and desirability as features that set social tipping apart from physical tipping. Fesenfeld et al. [8] studied the political conditions that make social tipping possible in climate policy. Swingedouw et al. [30] reviewed early-warning signs for tipping across physical, biological, and social systems, and noted that text evidence from the research record is still little used as an early signal. Machine-learning work on tipping has so far looked at early-warning signs in time-series data [31, 32], which sits alongside the text-based task taken up here rather than overlapping with it. The framework described below is the text-based counterpart to that time-series work and targets written STP evidence.

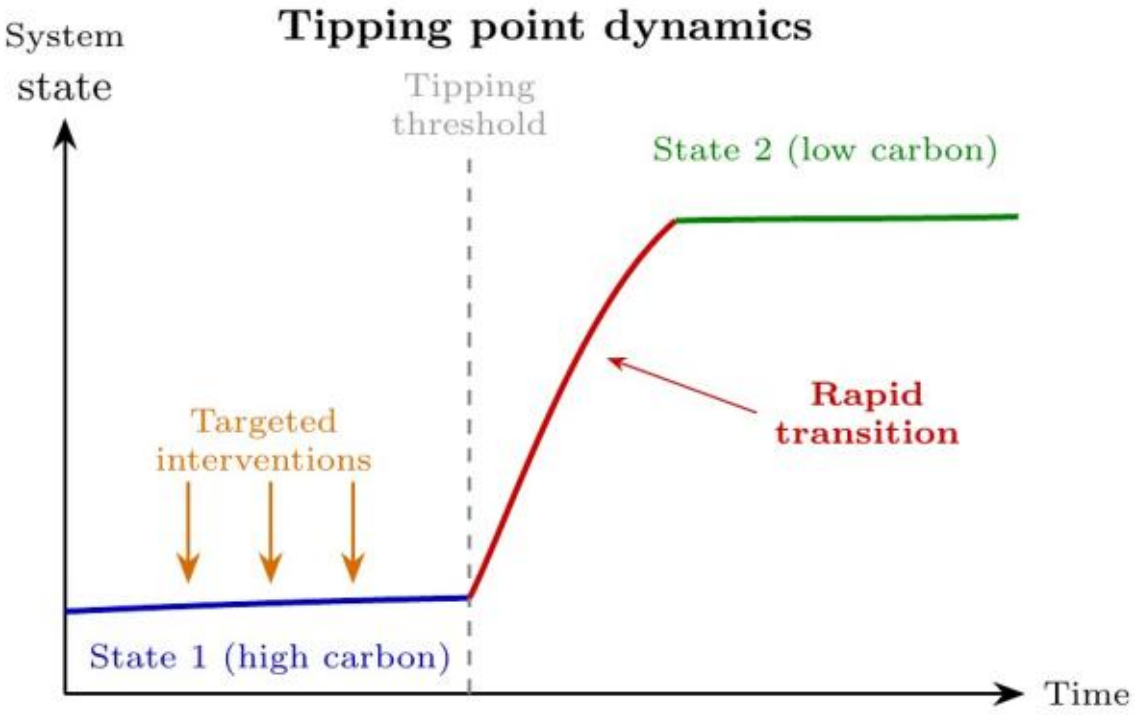

**Fig. 1** Tipping point dynamics. Targeted interventions push a social system toward its tipping threshold. Once that threshold is crossed, a rapid and self-sustaining shift carries the system from State 1 (high carbon) to State 2 (low carbon) without further intervention. The framework presented in this paper detects and structures the written evidence that documents this kind of transition in the scientific literature, policy reports, news, and online audio visual contents.

## 2.2 Machine learning for climate evidence synthesis

Several large-scale systems constitute the primary reference points against which the contribution of this work is assessed. Callaghan et al. [2] used a fine-tuned BERT classifier to find and map more than 100,000 studies of observed climate impacts, which showed that transformer classifiers can map evidence at a size and resolution hand review cannot reach. Callaghan et al. [4] grew this into a living map of climate policy papers covering 84,990 items, and Berrang-Ford et al. [14] built a similar map of 15,963 climate-and-health papers. Angin et al. [33] split sustainability reports into candidate paragraphs and matched them to disclosure labels, which is close in shape to the present pipeline but runs on company text and a fixed label set. Bingler et al. [16] fine-tuned ClimateBERT to sort disclosures in the style of the Task Force on Climate-related Financial Disclosures, and Stammbach et al. [15] reached high F1 on sentence-level detection of environmental claims. Ngee et al. [17] carried this over to environmental, social, and governance scoring from reporting data.

Recent reviews and benchmarks confirm that the passage-level concept task is the one left open. The survey by Volkanovska [3] covers climate language models and large language model systems and names detailed concept detection as the main unfilled need. Trajanov et al. [34] tested ChatGPT against ClimateBERT on five standard climate classification tasks and found that ClimateBERT did much better on short, domain-specific text, which supports tuning a domain model over using a large general model with no tuning. The ClimateEval benchmark of Kurfalı et al. [18] gives baselines across twenty-five climate text tasks, and West et al. [35] apply machine learning to track climate-policy progress in the European Green Deal. None of these systems sorts passage-level concepts in long-form climate documents, and none builds social tipping features into its training target.

The wider case for open, modular, well-documented pipelines over closed end-toend systems is supported by nearby work on shared infrastructure. Ouaknine et al. [36] built a shared catalogue of forest-monitoring datasets because closed pipelines do not let later teams reuse the parts. The system here follows the same idea. Every part can be replaced, the data is documented, and the rating rules are written out.

## 2.3 Text splitting, domain models, and data augmentation

Webersinke et al. [37] showed that training on 2 million climate paragraphs cuts masked-language error by about 48% and gives gains of 3 to 36% on later climate tasks. Later work confirmed the gain for domain models on broad climate text [15, 16]. The gain shrinks or reverses, though, on tasks that need a fine reading of meaning rather than topic spotting, and Trajanov et al. [34] report task-by-task gaps between ClimateBERT and ChatGPT. The same pattern shows up in adjacent work in this journal. Yan [38] reports that a standard two-stage pre-training-then-fine-tuning pipeline often fails to bridge the gap between a general pre-training and a specialised target domain, which is the failure mode that motivates the model comparison reported in this paper.

Data augmentation is a settled tool for text classification when labelled data is scarce [39]. Jain et al. [40] introduced failure-targeted augmentation in computer vision, using error analysis to design synthetic batches that hit a model's known weak spots. The strategy used here has the same shape, moved over to text classification, and the round-by-round record below gives the first published course for this kind of augmentation on a climate text task. Recent work on small fine-tuned language models for classification backs the view that a tuned encoder still holds its own against a much larger untuned model on narrow concept tasks [41, 42]. Closer in spirit, Yang [43] introduced a causal representation framework for few-shot text classification, arguing that surface statistical patterns alone do not give a robust signal when the training set is small. The augmentation strategy in this paper aims at the same problem from a different angle, by reshaping the training data to break specific patterns the model has latched onto.

Splitting a document into passages comes before any passage-level classification. Hearst [44] introduced lexical cohesion analysis for boundary finding, Choi [45] extended it with ranked sentence similarity, and Koshorek et al. [46] recast boundary finding as supervised labelling with a bidirectional LSTM. Topic-based splitting [47] gives clean topic passages but lower coverage when topics are spread unevenly. Wang et al. [48] report that retrieval performance depends heavily on splitting quality, which shows that splitting is not a neutral first step. DistilBERT [49] reaches about 97% of BERT's performance at 60% of the size, which makes it a practical choice for checking each adjacent sentence pair across a long document. Long-sequence modelling work in this journal [50] shows that scaling a single transformer to whole documents is itself an open problem, which strengthens the case for splitting before classification rather than feeding a long document to one model.

### 2.4 Research gap

Table 1 summarizes the main prior systems reported in literature for climate tipping point mapping, highlighting the dataset employed, methods, key findings and inherent limitation for tipping point monitoring. side by side and names, in its last column, the one reason each cannot detect social tipping points at the passage level without change. Three threads in the literature meet at a single open problem. The conceptual work sets out clear criteria for what a social tipping point is and how to tell it apart from physical tipping and from ordinary fast change [5, 6, 12, 22], but it stops short of turning those criteria into a measure that can be applied to text at scale, and several reviews name the cost of expert reading as the reason the field stays at the level of ideas [13, 19]. The machine-learning work maps climate evidence at a scale hand review cannot reach [2, 4, 14] or flags single-sentence claims and disclosures [15–17], but it works at the document level or the sentence level, not at the one-or-twoparagraph passage level where social tipping evidence sits. The recent ClimateEval benchmark carries no passage-level concept task [18], and the latest climate languagemodel survey names detailed concept detection as the main unfilled need [3]. Closer to the present work, recent papers in this journal have addressed the broader engineering challenge of adapting pre-trained language models to specialised domains and small labelled sets. Yang [43] introduced a causal representation framework for few-shot text classification. Yan [38] proposed a domain-adaptive fine-tuning framework that bridges the gap between general pre-training and a specialised target domain. Qi [50] addressed the linked problem of modelling long sequences for downstream tasks. Dantas et al. [51] reviewed model compression methods that matter when these pipelines have to run on the hardware most evidence teams have available. Each of these works addresses one part of the broader problem, but none combines passage-level segmentation, concept classification, and a structured criterion-based read-out into a single deployable system for applied evidence synthesis. That is the gap the present framework attempts to address.

## 3 The proposed framework

This section describes the proposed framework in full. Section 3.1 sets out the overall architecture and the flow of a document through the system. Section 3.2 describes the data used for training and evaluation. Section 3.3 explains the choice of each model. Sections 3.4 to 3.7 describe the four core processing stages in order. Section 3.8 describes the user interface and the deployment stack. Section 3.9 sets out how each stage was measured.

## 3.1 System overview

The framework is built as a modular pipeline. A document moves through four sequential processing stages, with intermediate outputs that can be inspected and replaced.

**Table 1** Earlier climate text and social tipping point work, with the present study in the last row. For each earlier system the "Tipping-point limit" column gives the one reason it cannot do passage-level tipping-point detection in long-form climate documents without change

| Study | Task | Method | Dataset | Key result | Tipping-point limitation |
|---|---|---|---|---|---|
| Callaghan et al. [2] | Map observed climate-impact studies | BERT-based evidence and attribution mapping | 100k+ climate-impact studies | Estimated 102,160 observed climate-impact studies | Whole-document evidence map, not passage-level tipping-point detection |
| Berrang-Ford et al. [14] | Map climate-and-health research | ML on abstracts, topic modelling, geoparsing | ¿16k relevant climate-health publications | Mapped evidence and gaps by topic and region | Whole-document or abstract map, health-only, no passage-level tipping-point labels |
| Webersinke et al. [37] | Climate text pre-training | Domain-adaptive pre-training (ClimateBERT) | 2M+ climate-related paragraphs | 48% MLM gain, 3.57–35.71% downstream error reduction | Climate text model, not tuned for tipping-point dynamics |
| Bingler et al. [16] | Climate-risk disclosure analysis | Fine-tuned ClimateBERT | TCFD-supporting company reports | Finds cheap talk and cherry-picking in voluntary disclosures | Company disclosure categories, not passage-level tipping-point evidence |
| Angin et al. [33] | SDG relevance classification | RoBERTa/BERT and classical ML | OSDG Community Dataset | Fine-tuned RoBERTa gives strong SDG classification performance | Fixed SDG labels, not tipping-point dynamics |
| Stammbach et al. [15] | Environmental-claim detection | Fine-tuned transformer models | 2,647 expert annotated company text examples | Transformer models reach above 82% F1, best test F1 about 84.9% | Sentence-level company claims, not passage-level tipping-point detection |
| Trajanov et al. [34] | ChatGPT vs ClimateBERT test | Untuned ChatGPT vs tuned climate NLP models | 5 climate text-classification tasks | ClimateBERT generally outperforms ChatGPT on classification | Short text classification, no tipping-point concept detection |
| Ngee et al. [17] | ESG scoring from GRI-linked reports | Sentence transformer similarity models | Company annual-report PDFs and GRI indicators | Best reported F1 is limited, about 0.46 | Company reports and GRI label matching only |
| Volkanovska [3] | Survey of climate language models | Review | N/A | Reviews climate LMs, systems, evaluation, data, access, and lifecycle issues | Survey only, no passage-level tipping-point detector |
| Kurfalı et al. [18] | 25-task climate text benchmark | LM Evaluation Harness | 13 datasets | Baselines across 25 climate NLP tasks | Benchmark lacks a passage-level social tipping-point task |
| Callaghan et al. [4] | Living map of climate policy literature | Transformer-based ML pipeline | 84,990 climate-policy papers | Maps policy papers by instrument, sector, and geography | Whole-document policy map, no passage-level tipping-point concepts |
| Leippold et al. [52] | Automated climate fact-checking | Climinator mediator-advocate LLM framework with sources | Climate claims with authoritative sources | Improves climate-claim checking with sourced LLM reasoning | Claim checking, not passage-level tipping-point detection |
| Ouaknine et al. [36] | Open data catalogue for forest ML | Shared dataset catalogue | 86 open-access forest datasets | Dynamic catalogue for forest-monitoring ML | Forest-monitoring data, not climate-text evidence extraction |
| **This study** | Passage-level social tipping point detection in climate documents | Pre-processing, DistilBERT segmentation, tuned RoBERTa classification, Mistral 7B rewriting, LLaMA 3.2 3B criterion-based rating, and Milvus retrieval | 90 expert-labelled positive passages, a 163-passage GPT-4.1 benchmark, and a 51-passage human check | DistilBERT splitter best of four (score 6.137), RoBERTa kappa 0.337 on the full set and 0.742 on clear cases | Fills the gap above: joins passage-level segmentation, concept classification, and a structured criterion read-out in one deployable system |

The four core stages are pre-processing and cleaning, segmentation into passages, classification of each passage, and post-processing of detected passages by rewriting and criterion-based rating. Figure 2 shows the overall system architecture, from raw data ingestion through all processing stages to the downstream retrieval systems that consume the indexed results. Figure 3 gives a detailed block diagram of the internal pipeline stages.

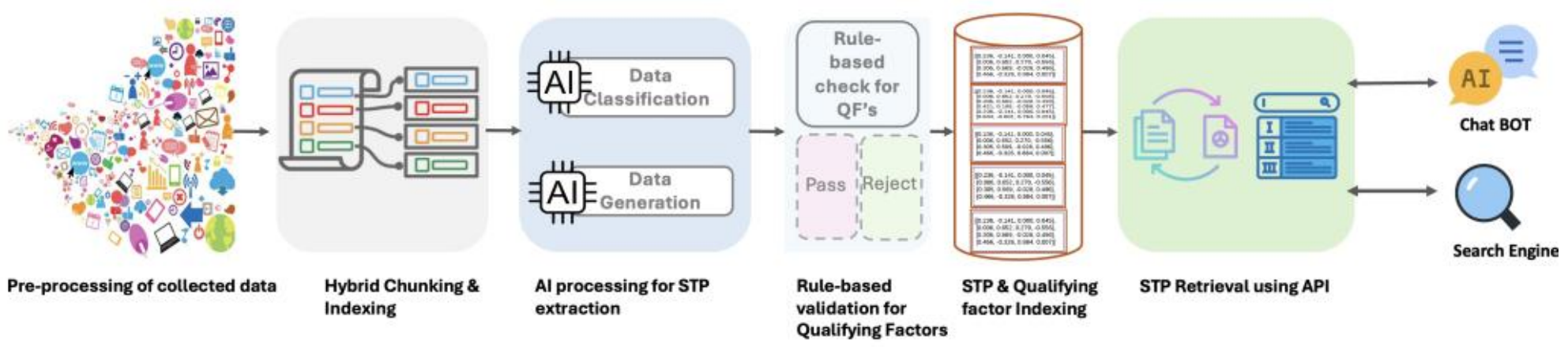

**Fig. 2** System architecture of the proposed STP detection and retrieval framework. Raw source data is pre-processed and segmented through hybrid chunking and indexing. AI classification identifies candidate STP passages while a parallel generation step produces qualifying factor assessments. A rule-based check filters each passage against the qualifying factor criteria. Those that do not pass are rejected. Accepted passages and their qualifying factor scores are stored in Milvus. A retrieval API exposes the indexed store to two downstream systems: the conversational climate assistant, and the document search interface.

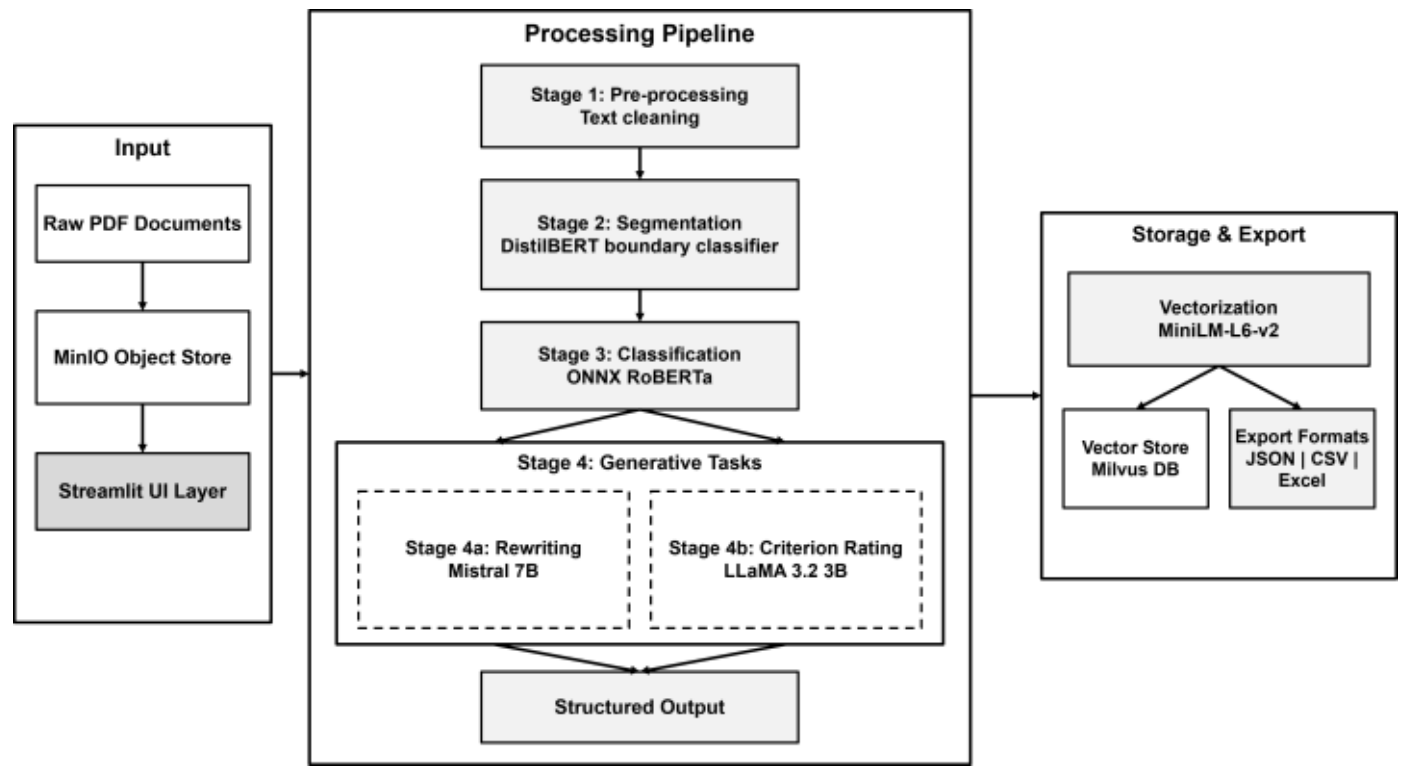

**Fig. 3** Detailed block diagram of the proposed STP detection framework. Documents enter the system as raw PDFs, are stored in a MinIO object store, and are submitted for processing through a Streamlit user interface. The core processing pipeline has four sequential stages. Stage 1 cleans raw text using Unstructured, ftfy, NLTK, WordNinja, and a reference removal step. Stage 2 segments cleaned text into passages with a DistilBERT boundary classifier, sentence-pair scoring, and passage assembly. Stage 3 classifies each passage as STP or non-STP using an ONNX-compiled RoBERTa model Stage 4 applies two parallel generative tasks to detected passages: Stage 4a rewrites each passage with Mistral 7B via vLLM and Stage 4b rates each passage against five social tipping point criteria with LLaMA 3.2 3B model via vLLM. Processed passages are vectorised with MiniLM-L6-v2 and stored in a Milvus vector database. Results can be exported as JSON, CSV, or Excel.

Each stage in Figure 3 reads from and writes to a shared infrastructure layer. Raw PDFs are stored in a deployed instance of MinIO, an S3-compatible object store. The classifier runs as an ONNX export for fast inference. Detected passages, their rewritten forms, and their criterion ratings are embedded with MiniLM-L6-v2 and stored in Milvus, where they can be queried by the downstream systems shown in Figure 2. A Streamlit application wraps the whole workflow and exposes each stage as its own configurable panel. Section 3.8 describes this interface in more detail.

## 3.2 Building the dataset

*Training data.*

The positive examples are 90 passages labelled by hand across three co-design activities run with project partners. A sensemaking workshop titled *STPs in practice: a cross-role dialogue on climate adaptation* brought together 24 consortium members, among them policymakers, educators, youth workers, and journalists, around a shared working understanding of social tipping points. A second activity turned that understanding into training data through collaborative passage labelling. The third co-designed the visual interface used to browse and review detected passages. The labelled set covers floods, air pollution, over-tourism, and plastic pollution, each marked against five social tipping point qualifying factors:

1. felt social consequences
2. shared awareness
3. shared understanding of causes and effects
4. demand for changes in habits
5. demand for political action

Ninety negative examples come from the ClimateBERT climate-specificity database, which gives climate text without tipping-point content. Drawing the negatives from the same domain forces the classifier to tell tipping points apart from ordinary climate writing rather than from off-topic text, which makes the task harder and the results more convincing.

The resulting training dataset is relatively small compared with conventional NLP benchmarks. However, social tipping point identification requires substantial domain expertise and careful interpretation of contextual evidence, making large-scale manual annotation costly and time-intensive. The objective of this work was therefore not to construct a large benchmark dataset, but to investigate whether a modular STP detection framework could be developed and iteratively improved in the few-shot, lowresource conditions typical of emerging interdisciplinary research. This setting reflects practical constraints faced by many climate and sustainability research initiatives where expert-labelled data are limited.

*Test data and confidence levels.*

For the held-out evaluation, GPT-4.1 generated initial labels for 163 passages extracted from documents that were not used during training. These annotations were subsequently reviewed and curated by researchers using the five social tipping point (STP) criteria, producing the benchmark dataset used for model evaluation. Each passage was also assigned a confidence level (LOW, MEDIUM, or HIGH) reflecting the strength of evidence supporting the assigned label.

Since expert-labelled evaluation data for social tipping points are scarce and costly to produce, this semi-automated annotation approach was used to create a benchmark of sufficient size for comparative evaluation. As shown in Figure 4, the resulting dataset consisted of 72 HIGH-confidence passages, 89 MEDIUM-confidence passages, and 2 LOW-confidence passages. The LOW-confidence passages were excluded from further analysis.

Results are reported on two evaluation subsets. Set A contains only HIGHconfidence passages (72 samples: 44 Non-STP and 28 STP) and is used to assess model performance on clear and unambiguous cases. Set B contains both HIGH- and MEDIUM-confidence passages (161 samples: 117 Non-STP and 44 STP) and provides a more challenging evaluation that includes borderline cases. Comparing performance across the two sets enables assessment of model robustness under different levels of annotation certainty. Figure 4 shows how the sets were built.

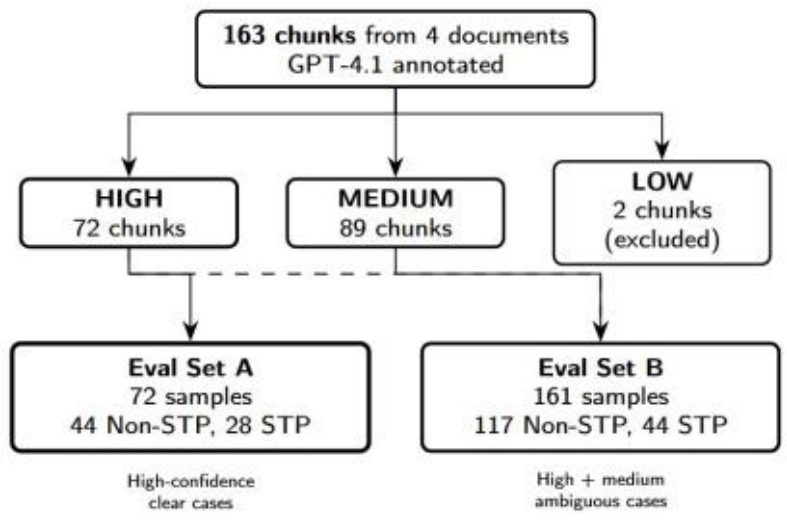

**Fig. 4** Construction of the evaluation benchmark. GPT-4.1 generated initial labels for 163 passages, which were subsequently reviewed and curated by researchers and assigned confidence levels. Two LOW-confidence passages were excluded. Set A (HIGH only; 72 passages) evaluates performance on clear cases, while Set B (HIGH + MEDIUM; 161 passages) evaluates performance on both clear and borderline cases.

## 3.3 Choosing the models

Three factors guided the choice of models: fit to climate text, available hardware for deployment (NVIDIA RTX 4070 GPUs), and fit for tuning on a small labelled set.

ClimateBERT [37] was the baseline trained on climate data and thus the natural point of comparison. If a model already tuned to climate data cannot find tipping points well, that fixes how hard the task is and shows why more tuning is needed. RoBERTa [53] was the primary classification model, as it performs well on sentence and paragraph classification with minimal tuning data. Its general representation is less likely to bias that would compoundhat would stack on top of a small training signal. The tuned RoBERTa is exported to ONNX format and shipped with the system, which keeps inference fast on commodity hardware.

Mistral 7B was chosen for the rewriting step because it follows instructions well at a size that fits in memory without compression. For the rating step, a LLaMA 3.2 3B model is used. The smaller model with task-specific tuning gave more consistent structured output on the criterion rating task than larger untuned models, while staying within the compute budget of the deployed system. LLaMA 3.2 3B [54] was also the untuned classification baseline, which gives a comparison between large untuned models and the tuned-encoder approach at a similar cost. MiniLM-L6-v2 [55] was used to make the embeddings because it balances speed and meaning at 384 dimensions.

### 3.4 Pre-processing: text extraction and cleaning

The Unstructured framework reads each PDF into labelled parts: body text, titles, list items, tables, and figures. Tables, figures, repeated or decorative page elements, and one-word fragments are dropped. A five-step cleaning pass is then run on each text part. The ftfy library fixes broken characters such as the common fi ligature corruption. An NLTK chunker shields named terms (people, places, organisations) so they are not split wrongly. A set of thirteen regular expressions fixes spacing across common patterns such as missing spaces after commas or full stops. The WordNinja tool splits run-together words such as "soybeansfromArgentina" using Wikipedia frequencies and dynamic programming, with the result checked against an NLTK word list to guard against false splits. A final pass standardises white space. Reference lists are then found and removed by a multi-signal detector that looks for reference section headings, copyright and licence notices, numbered citation lists, author declaration blocks, URL-heavy paragraphs, and journal citation patterns. Once any signal fires, all text from that point to the end of the document is removed.

PDF text often breaks sentences across lines or pages. A rule based on transition words finds and rejoins these broken pieces so that each sentence reads as one. The output of pre-processing is a clean stream of whole sentences, ready for segmentation.

### 3.5 Segmentation: splitting documents into passages

Each source type poses its own segmentation problem. Tipping-point passages are short, often one or two paragraphs, and sit inside longer sections of related but ordinary climate content. Token-based and fixed-window methods break those passages up, which fits the general findings of Hearst [44] and the retrieval-specific findings of Wang et al. [48].

The cleaned, rejoined text is split into sentences and passed to a DistilBERT boundary classifier, which decides for each pair of neighbouring sentences whether the two belong to the same passage. This puts the boundary question directly to the model rather than guessing at it with a fixed window or a topic cluster, and it carries the supervised boundary idea of Koshorek et al. [46] over to a transformer. The passages that come out hold their topic together across the many layouts of these documents. Figure 5 shows how the methods differ on the same document.

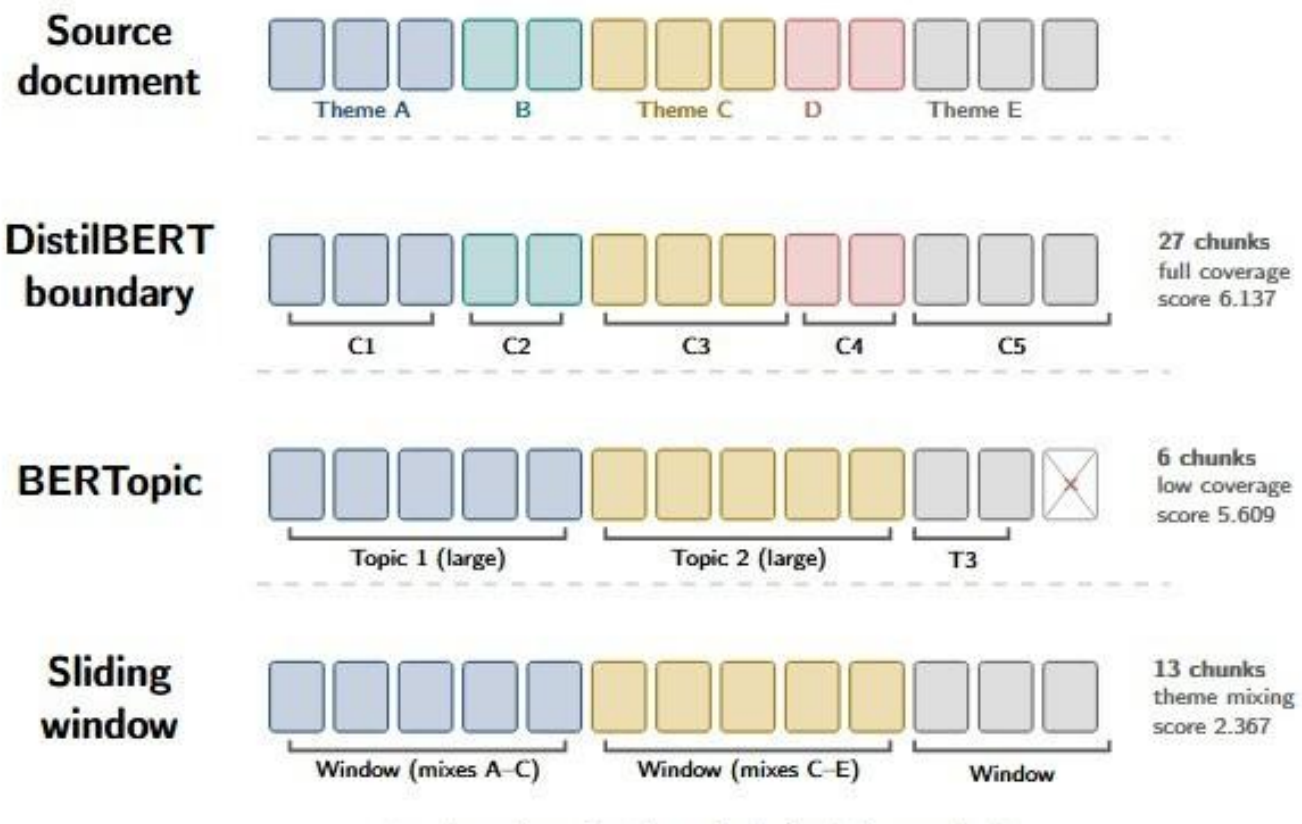

**Fig. 5** How three splitting methods treat the same 13-part document. Fill colour shows the true topic of each part. DistilBERT boundary finding respects topic edges and assigns every part, giving small single-topic passages. BERTopic groups parts across topics into a few large clusters and leaves some parts out. The fixed window cuts at set points, so single passages cross topic edges and no topic is kept whole

## 3.6 Classification: sorting passages

### 3.6.1 Training setup

RoBERTa and ClimateBERT were tuned for three epochs with standard settings for two-class text classification, using the HuggingFace Transformers library [56]. Training ran on NVIDIA RTX 4070. Cross-validation held overfitting in check on the small labelled set. The final version from each round was picked based on balanced performance across tipping-point F1, Cohen's kappa, and steadiness across confidence levels, with class balance valued over plain accuracy on the majority class. Once the best classifier was selected, it was exported to ONNX format for fast inference inside the deployed system.

### 3.6.2 Targeted augmentation

Four rounds of augmentation followed classification errors found through repeated error analysis on the training set. The design follows the failure-targeted idea of Jain et al. [40], moved from vision to text. The training set kept a 70:30 mix of real to synthetic examples across all rounds, in line with Wei and Zou [39]. The fixed mix held the shape of real tipping-point text while still fixing the target weak spots. Figure 6 shows the error each round went after and the synthetic data it added.

Round 1 (V0.2) went after a length bias. The base model linked longer passages to the tipping-point label. Ninety synthetic examples balanced this, with 45 short positive passages and 45 long negative ones that broke the tie between length and class. Round 2 (V0.3) went after a mix-up between the fast onset of tipping points and the slow change of long-run climate processes. Thirty synthetic examples made the contrast plain at the sentence level. Round 3 (V0.4) tried to fix a mix-up between theory passages that discuss tipping points and passages that document real ones, where the model struggled because both use the same vocabulary in a different register. Sixty-five examples were generated and tested, but they pushed the model into an over-cautious state, so the V0.4 batch was not included in the deployed system. Round 4 (V0.5) added ten negative examples drawn from reference lists, which early tests showed were wrongly marked as tipping points because they were dense with tipping-point terms.

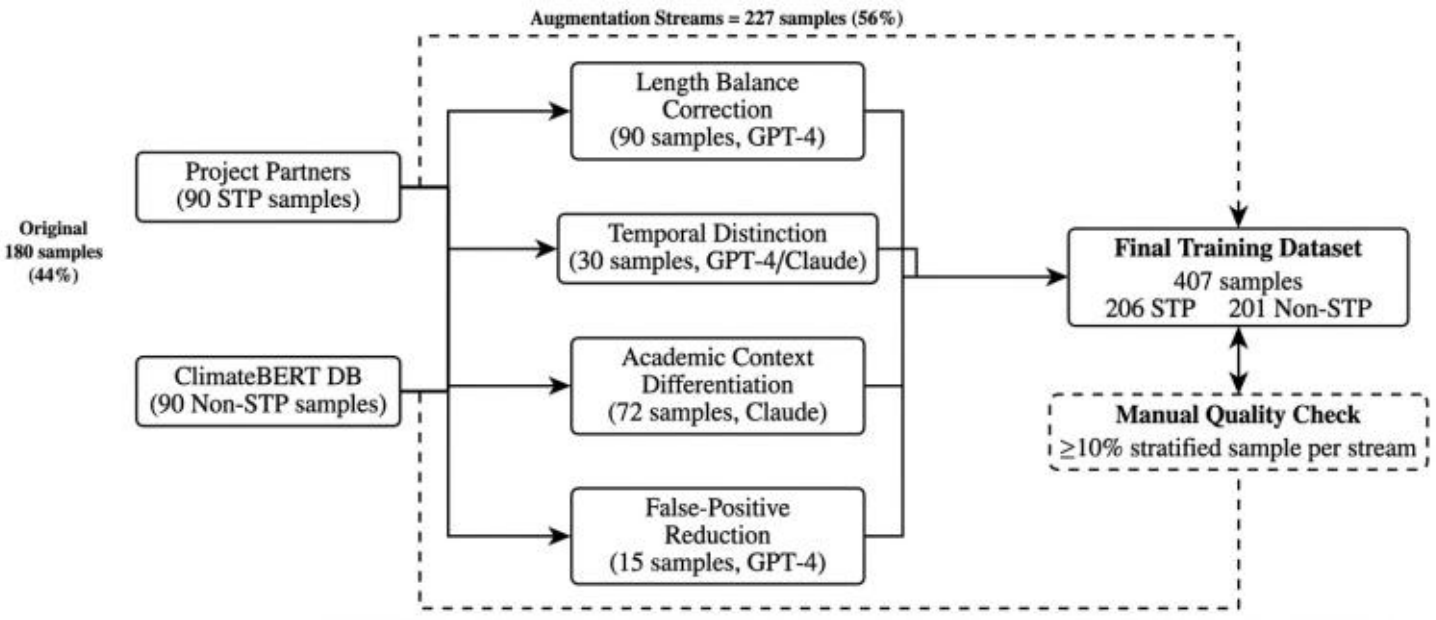

**Fig. 6** The augmentation plan and the make-up of the training set across the four rounds (V0.2 to V0.5). Dashed arrows show real expert-labelled examples flowing into the training set. Solid arrows show the four synthetic streams, each made by prompt-guided models (GPT-4 and Claude Sonnet) to fix one known classification error. The quality step required a stratified hand check of at least 10% of each stream before it went in. The set kept a fixed 70:30 mix of real to synthetic examples across rounds

## 3.7 Post-processing: rewriting and rating

Two post-processing steps turn raw detected passages into a structured, readable evidence record. A rewriting step makes each passage clear without changing its meaning, and a rating step scores it against the five social tipping point criteria.

### 3.7.1 Rewriting passages for clarity

Detected passages are often fragmented and dense; they may begin or end abruptly and may not form coherent, self-contained pieces of information when read in isolation, which limits their usefulness for a policy reader. An automated rewriting step addresses this. A structured prompt turns each detected passage into a stand-alone, readable note while keeping the facts and meaning. The prompt tells the Mistral 7B model to keep every factual claim, reorder sentences for clarity, and add nothing that is not in the source. This grounding rule follows the design of Leippold et al. [52].

### 3.7.2 Rating the five criteria

Each rewritten passage is rated against five qualifying factors from the social tipping point literature [5, 6]: felt social consequences, shared awareness, shared understanding of causes and effects, demand for changes in habits, and demand for political action. The rating step uses LLaMA 3.2 3B model, trained to produce structured outputs that name only the criteria actually supported by the text, together with a confidence level and a short evidence note. Each rating is one of *Strong*, *Moderate*, *Weak* or *Not evident*. A passage clears the quality bar if at least one of the five criteria are rated Strong, Moderate or Week. The rewriting and rating steps together form a structured tipping-point summary fit for both expert review and policy use.

## 3.8 User interface and deployment

The framework configuration exposed to the user through a Streamlit application.The application is structured such that each pipeline stage can be configured independently through a dedicated sidebar panel and executed from a corresponding tab in the main interface. Figure 7 shows the eight collapsible configuration sections of the sidebar.

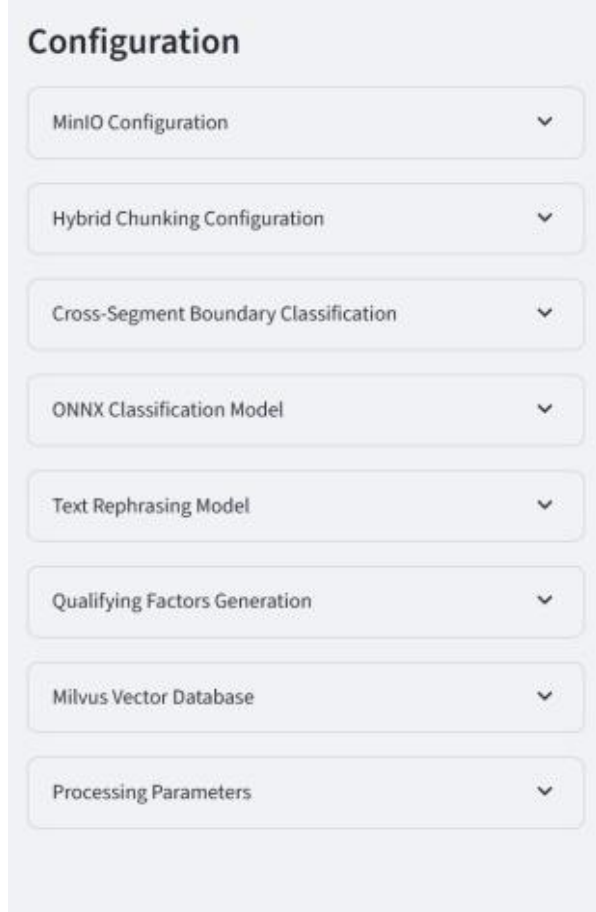

**Fig. 7** The configuration sidebar of the Streamlit application. Each collapsible section exposes the parameters of one stage of the pipeline: MinIO endpoint and credentials, hybrid chunking parameters (strategy, layout model, table inference, character limits, overlap), the cross-segment boundary classification threshold, the ONNX classification model, the text rephrasing model, the qualifying-factors generator, the Milvus vector store endpoint, and general processing parameters. Each setting is wired to a single stage of the pipeline, which lets a user swap in a new model or tune a single threshold without touching the rest of the system

The MinIO panel sets the endpoint, access key, and secret key for the object store. The hybrid chunking panel sets the document layout strategy (hi-res, OCR-fallback, or fast), the visual layout model, whether to infer table structure or include element coordinates, the character limit per chunk, the character overlap between chunks, and the merge threshold for short fragments. The cross-segment boundary classification panel sets the threshold used by the DistilBERT splitter to decide whether two neighbouring sentences sit inside the same passage. The ONNX classification model panel reads available models from an onnx-exports directory at start-up and lets the user pick one, which makes it straightforward to compare a freshly tuned classifier with an earlier release without restarting the server. The text rephrasing and qualifying-factor panels point at the Mistral and LLaMA endpoints respectively, with their own temperature and token-limit fields. The Milvus panel sets the vector store endpoint and the target collection name. The processing-parameters panel groups batch sizes and concurrency settings.

The main interface is structured as an ordered sequence of tabs, each corresponding to a distinct stage of the pipeline. The first tab connects to MinIO and lists the available PDF files. The chunking tab runs the segmentation step and lets the user save or load chunks for re-use, which speeds up experimentation when only the downstream models are being changed. The classification tab applies the ONNX classifier to the loaded chunks. The qualifying-factors tab runs the LLaMA rating model on the passages flagged by the classifier. The results tab gives a preview of the classification and rating outputs, with a per-passage view. Figure 8 shows the processing summary for a complete run. The export tab writes the results out in JSON, CSV, or Excel form, with a field selector that lets the user choose which columns to include. Detected passages, their rewritten versions, and their criterion ratings are also embedded and pushed into Milvus, where they can be queried by retrieval-augmented systems built on top of the same store.

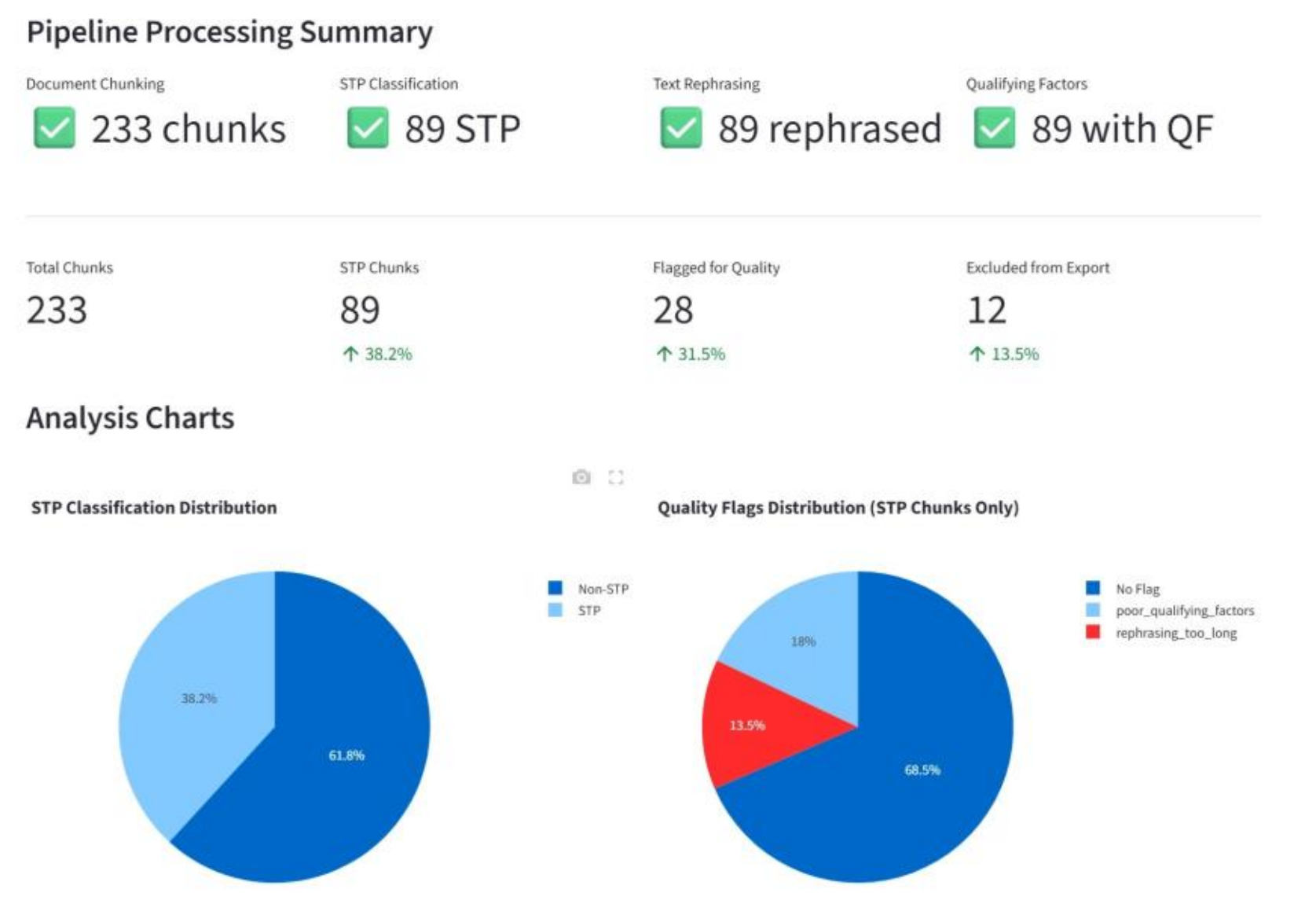

**Fig. 8** Processing summary view of the Streamlit application after a complete pipeline run. The header row tracks each stage: 233 passages produced by segmentation, 89 detected as social tipping points (38.2%), all 89 rephrased, and all 89 rated against the five criteria. The left pie chart shows the overall STP-to-non-STP split across the knowledge base. The right chart breaks the 89 tipping-point passages by quality flag: 68.5% cleared with no flag, 18% were flagged for poor qualifying factors, and 13.5% were flagged for rephrasing that grew too long. The summary view is generated automatically at the end of each run and can be exported alongside the structured passage results

The system was designed so that the four model components can be replaced on their own. A new boundary splitter, a new tuned classifier, a new rephraser, or a new criterion rater can be dropped in by changing one line in the relevant configuration panel and pointing it at the new endpoint or ONNX file. Storing the passages, the classifier outputs, and the rated passages in a shared object store and a shared vector database means that downstream consumers, including retrieval-augmented question answering and policy-facing dashboards, can be built without having to re-run the upstream stages.

## 3.9 Performance measurement

*Splitting.*

Splitting was scored on nine measures, each scaled to [0,1] and added into one combined score: silhouette score [57], Calinski–Harabasz index [58], inverted Davies– Bouldin index [59], within-passage cosine similarity, topic coherence, coverage, semantic density, inverted Gini coefficient for size balance, and page continuity. For the nine normalised component scores $s_1,...,s_9$ the combined score is their sum,

$$S = \sum_{k=1}^{9} \tilde{s}_k, \qquad \tilde{s}_k \in [0,1], \tag{1}$$

All component scores were normalised to the range [0,1], except coverage, which could slightly exceed 1.0 under the assignment-reward scaling used to favour complete document coverage. Also a method scores well only when it performs consistently across multiple aspects of segmentation quality rather than excelling on a single metric. Since no prior work establishes the relative importance of these criteria for STP passage extraction, equal weighting was adopted after normalisation to avoid introducing subjective bias into the evaluation. Among the components, the three standard clustering indices are defined as follows:

The silhouette score for a part with mean within-passage distance $a$ and mean nearest-other-passage distance $b$ is

$$s = \frac{b-a}{\max(a,b)}. \tag{2}$$

The Davies–Bouldin index, inverted so that higher is better, averages over $K$ passages the worst-case ratio of within-passage scatter $\sigma_i$ to between-passage distance $d_{ij}$,

$$\mathrm{DB} = \frac{1}{K}\sum_{i=1}^{K} \max_{j \neq i} \frac{\sigma_i + \sigma_j}{d_{ij}}. \tag{3}$$

The Calinski–Harabasz index relates between-passage dispersion $B_K$ to within-passage dispersion $W_K$ for $N$ parts,

$$\mathrm{CH} = \frac{B_K}{W_K} \cdot \frac{N-K}{K-1}. \tag{4}$$

The remaining components are domain measures for this task. Within-passage cosine similarity is the mean pairwise similarity of part embeddings inside a passage, coverage is the share of document parts assigned to a passage, semantic density rewards passages that stay on one topic, the inverted Gini coefficient rewards even passage sizes, and page continuity rewards passages whose parts sit close together in the document.

*Classification.*

Classification was scored on accuracy, precision, recall, F1 per class, macro and weighted F1, and Cohen's kappa [60]. Writing *TP*, *TN*, *FP*, and *FN* for true positives, true negatives, false positives, and false negatives, the per-class measures are

$$\text{Precision} = \frac{TP}{TP+FP}, \qquad \text{Recall} = \frac{TP}{TP+FN}, \tag{5}$$

$$\text{F1} = 2 \cdot \frac{\text{Precision} \cdot \text{Recall}}{\text{Precision} + \text{Recall}}, \qquad \text{Accuracy} = \frac{TP+TN}{TP+TN+FP+FN}. \tag{6}$$

Macro F1 is the unweighted mean of the per-class F1 scores, and weighted F1 averages them in proportion to class support. Cohen's kappa adjusts the observed agreement $p_o$ for the agreement $p_e$ expected by chance,

$$\kappa = \frac{p_o - p_e}{1 - p_e}. \tag{7}$$

Kappa measures agreement between the model and the labels after the agreement expected by chance is removed, and it is the main measure used to compare augmentation rounds because the test sets hold more of one class than the other.

*Rewriting.*

Rewriting was scored on mean cosine similarity between source and rewritten passage embeddings, Jaccard similarity for word overlap, and lexical richness as the type-token ratio. For embeddings **a** of the source and **b** of the rewrite, and for word sets *A* and *B* of the two passages, these are

$$\cos(\mathbf{a},\mathbf{b}) = \frac{\mathbf{a}\cdot\mathbf{b}}{\|\mathbf{a}\|\|\mathbf{b}\|}, \qquad J(A,B) = \frac{|A \cap B|}{|A \cup B|}, \tag{8}$$

number of unique words

TTR = 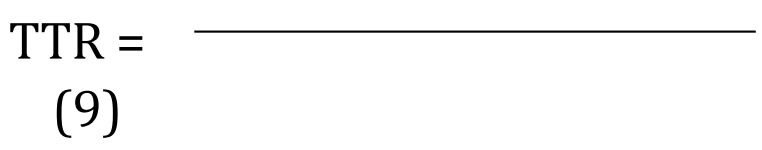 . (9)

total number of words

High cosine similarity together with low Jaccard similarity is the mark of a real rewrite rather than a near-copy.

# 4 Results

## 4.1 Document chunking performance

DistilBERT boundary splitting reached the top combined score (6.137) of the four methods on a 149-part climate document. Table 2 gives the full result. The document held 104 body-text parts, 42 titles, and 3 list items.

DistilBERT generated 27 passages, with a mean size of 2.9 parts, within-passage similarity of 0.562, and coverage of 1.149. Coverage above 1.0 reflects a scaling rule that rewards near-complete assignment, and DistilBERT placed all 149 parts with no gaps. BERTopic generated 6 passages of mean size 9.0 and a combined score of 5.609. Its coherence (0.474) was the best of the four, as expected for large singletopic passages, but its coverage (0.806) was lower because parts that did not fit any topic were left out. The semantic-similarity method generated 10 passages with the best within-passage similarity (0.661) but the lowest coverage (0.612), which shows that it forms tight groups at the cost of leaving text out. The fixed window had the lowest combined score (2.367) and the lowest within-passage similarity (0.428), which confirms that these documents do not switch topic at set points.

**Table 2** Splitting method results on a 149-part climate document. The combined "Score" is the sum of nine measures each scaled to 0–1. Coverage above 1.0 for DistilBERT reflects the scaling rule that rewards near-complete assignment. The best value in each column is in bold

| Method | Score | Passages | Avg size | Within-sim | Coherence | Coverage | Density | Gini |
|---|---|---|---|---|---|---|---|---|
| DistilBERT boundary | **6.137** | 27 | 2.9 | 0.562 | 0.314 | **1.149** | **0.692** | 0.231 |
| BERTopic | 5.609 | 6 | 9.0 | 0.548 | **0.474** | 0.806 | 0.620 | **0.185** |
| Semantic similarity | 4.457 | 10 | 4.1 | **0.661** | 0.327 | 0.612 | 0.679 | 0.290 |
| Fixed window | 2.367 | 13 | 5.0 | 0.428 | 0.353 | 0.970 | 0.607 | 0.230 |

DistilBERT generated smaller passages on average, which fits the fact that tippingpoint passages are one or two paragraphs set inside longer sections. The larger passages from BERTopic and the semantic-similarity method risk mixing tipping-point and ordinary content in one passage, which then weakens the classification step. Figure 9 shows the same comparison in the application's own evaluation dashboard, with the combined score broken down across each metric category. On the joint test of coherence and coverage the DistilBERT boundary splitter came out ahead and was selected as the segmentation component of the deployed system.

## 4.2 Classification performance

The tuned RoBERTa reached kappa 0.337 on the combined Set B (71.4% accuracy) and kappa 0.742 on the high-confidence Set A (87.5% accuracy), ahead of the climatetrained ClimateBERT and the untuned Mistral and LLaMA on balanced detection. The score also rose overall across the four augmentation rounds, from a base of 0.284 to 0.337, with a short dip at the third round. Tables 3 and 4 give the full results on Set B (161 passages: 117 non-tipping, 44 tipping) and Set A (72 passages: 44 non-tipping, 28 tipping).

On Set B, RoBERTa V0.5 reached 71.4% accuracy, tipping-point precision 0.48, tipping-point recall 0.61, and kappa 0.337. The base model reached 61.5%. Each round raised accuracy: V0.2 65.8%, V0.3 69.6%, V0.4 70.2%, V0.5 71.4%. Among the untuned models, LLaMA 3.2 3B reached 74.5% accuracy but a tipping-point F1 of only 0.25, because its recall was 0.16. It marked nearly everything as non-tipping, which looks accurate but misses the class that matters. Mistral 8B and ClimateBERT did the reverse, with recall near 1.00 but precision near 0.31, which means they marked almost everything as a tipping point. Augmented ClimateBERT V0.2 did better than the base, at 58.4% accuracy and kappa 0.269, but stayed below every RoBERTa version.

On the cleaner Set A, every model did better. RoBERTa V0.2 was best, with 87.5% accuracy, tipping-point F1 0.85, and kappa 0.742. The base model was close, at 84.7%. The gap between V0.2 and the later rounds on Set A (V0.3 kappa 0.480, V0.4 0.548, V0.5 0.613) mirrors the dip on Set B and shows that the later rounds, which were tuned for hard cases, gave back a little on clear ones. The 16-point accuracy gap between Set A and Set B for the top models (87.5% against 71.4%) points to where the difficulty lies. The models are reliable when the text clearly is or clearly is not a tipping point, and they struggle on passages where even human raters are unsure. Cohen's kappa was included to account for agreement beyond chance and is particularly informative in imbalanced classification settings. According to the commonly used interpretation of Landis and Koch, values between 0.21–0.40 indicate fair agreement, 0.41–0.60 moderate agreement, and values above 0.60 substantial agreement. The kappa values obtained on Set A (0.689–0.742 for the best-performing models) therefore indicate substantial agreement between model predictions and benchmark labels, whereas the lower values observed on Set B reflect the greater ambiguity introduced by the medium-confidence passages. The deployed system ships with both V0.2 and V0.5 available through the ONNX classifier panel of the configuration interface, since V0.2 is the right choice for clear-case workloads and V0.5 is the right choice for mixed-quality real-world inputs.

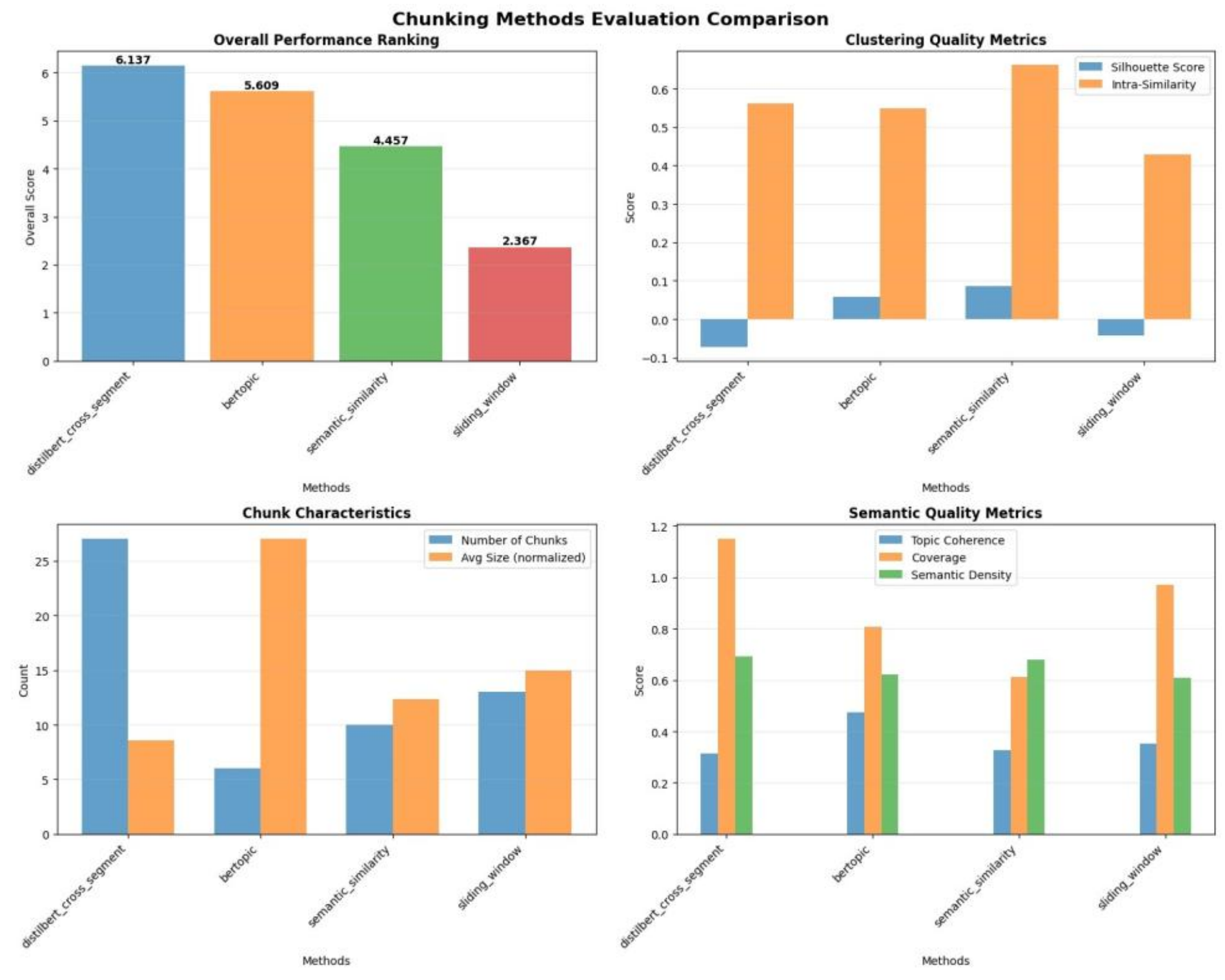

**Fig. 9** The application's evaluation dashboard for the four splitting methods on the same 149-part document used in Table 2. Top left: the combined nine-metric score, with DistilBERT boundary splitting at 6.137, BERTopic at 5.609, semantic similarity at 4.457, and the fixed window at 2.367. Top right: clustering quality, where the silhouette score and the within-passage cosine similarity track each other across methods. Bottom left: passage characteristics, where DistilBERT made many small passages while BERTopic made a few very large ones. Bottom right: semantic quality scored across topic coherence, coverage, and semantic density. The dashboard view supports the conclusion drawn from the table. No single method wins on every measure, and the combined score rewards methods that perform well across categories rather than ones that do well on only one. The application produces this view automatically in its chunking-evaluation tab, which lets a user reproduce the comparison on any uploaded document

**Table 3** Classification results on Set B (161 passages: 117 non-tipping, 44 tipping). Best values are in bold

| Model | Acc | Non-tip P | Non-tip R | Tip P | Tip R | Tip F1 | Macro F1 | Kappa |
|---|---|---|---|---|---|---|---|---|
| RoBERTa (base) | 0.615 | 0.91 | 0.52 | 0.40 | 0.86 | 0.55 | 0.61 | 0.284 |
| RoBERTa V0.2 | 0.658 | 0.91 | 0.59 | 0.44 | 0.84 | 0.57 | 0.64 | 0.334 |
| RoBERTa V0.3 | 0.696 | 0.81 | 0.76 | 0.45 | 0.52 | 0.48 | 0.63 | 0.270 |
| RoBERTa V0.4 | 0.702 | 0.83 | 0.74 | 0.46 | 0.57 | 0.51 | 0.65 | 0.305 |
| **RoBERTa V0.5** | **0.714** | **0.84** | 0.75 | **0.48** | 0.61 | **0.54** | **0.67** | **0.337** |
| LLaMA 3.2 3B | 0.745 | 0.75 | 0.97 | 0.64 | 0.16 | 0.25 | 0.55 | 0.163 |
| Mistral 8B | 0.398 | 1.00 | 0.17 | 0.31 | 1.00 | 0.48 | 0.38 | 0.101 |
| ClimateBERT (base) | 0.385 | 1.00 | 0.15 | 0.31 | 1.00 | 0.47 | 0.37 | 0.090 |
| ClimateBERT V0.2 | 0.584 | 0.95 | 0.45 | 0.39 | 0.93 | 0.55 | 0.58 | 0.269 |

**Table 4** Classification results on Set A (72 high-confidence passages: 44 non-tipping, 28 tipping). Best values are in bold

| Model | Acc | Non-tip P | Non-tip R | Tip P | Tip R | Tip F1 | Macro F1 | Kappa |
|---|---|---|---|---|---|---|---|---|
| RoBERTa (base) | 0.847 | 0.92 | 0.82 | 0.76 | 0.89 | 0.82 | 0.84 | 0.689 |
| **RoBERTa V0.2** | **0.875** | **0.93** | 0.86 | 0.81 | 0.89 | **0.85** | **0.87** | **0.742** |
| RoBERTa V0.3 | 0.764 | 0.76 | 0.89 | 0.76 | 0.57 | 0.65 | 0.74 | 0.480 |
| RoBERTa V0.4 | 0.806 | 0.80 | 0.88 | 0.78 | 0.64 | 0.70 | 0.78 | 0.548 |
| RoBERTa V0.5 | 0.819 | 0.83 | 0.89 | 0.80 | 0.71 | 0.75 | 0.81 | 0.613 |
| LLaMA 3.2 3B | 0.708 | 0.68 | 1.00 | 1.00 | 0.25 | 0.40 | 0.60 | 0.289 |
| Mistral 8B | 0.639 | 1.00 | 0.41 | 0.52 | 1.00 | 0.68 | 0.63 | 0.350 |
| ClimateBERT (base) | 0.583 | 1.00 | 0.32 | 0.48 | 1.00 | 0.65 | 0.57 | 0.266 |
| ClimateBERT V0.2 | 0.792 | 0.91 | 0.73 | 0.68 | 0.89 | 0.77 | 0.79 | 0.586 |

Prior to any augmentation, both base ClimateBERT and untuned Mistral 8B default to a degenerate positive prediction, assigning the STP label to all passages (precision 0.31, recall 1.00). Augmentation raises ClimateBERT V0.2 to a Cohen's kappa of 0.269. The general-domain RoBERTa V0.5, trained under the same augmentation protocol, yields a Cohen's kappa of 0.337. On the high-confidence subset (Set A), RoBERTa V0.2 achieves a Cohen's kappa of 0.742. The RoBERTa results sit in the balanced part of the precision-recall space, while the others sit at the edges, as Figure 10 shows. The pattern matches the failure modes that Trajanov et al. [34] report for untuned models on domain tasks.

The augmentation course on Set B is itself informative. The score rose overall (base 0.284 to V0.5 0.337) but not at every step, with a dip at V0.3 (0.270) before it recovered at V0.4 (0.305) and V0.5 (0.337). The dip came from a class-balance shift caused by the temporal-confusion fix, which sharpened the line between slow and fast change but pushed some borderline positives over to the negative side. Round 4 corrected this without bringing back the first bias. The Set A picture adds a sharper version of the same effect. V0.2 is the peak on Set A (kappa 0.742), and the later rounds, tuned for hard cases, gave back some ground on clear ones. Each round moves the decision line in a way that helps one group at the cost of the other, and the four rounds together give a better but uneven result.

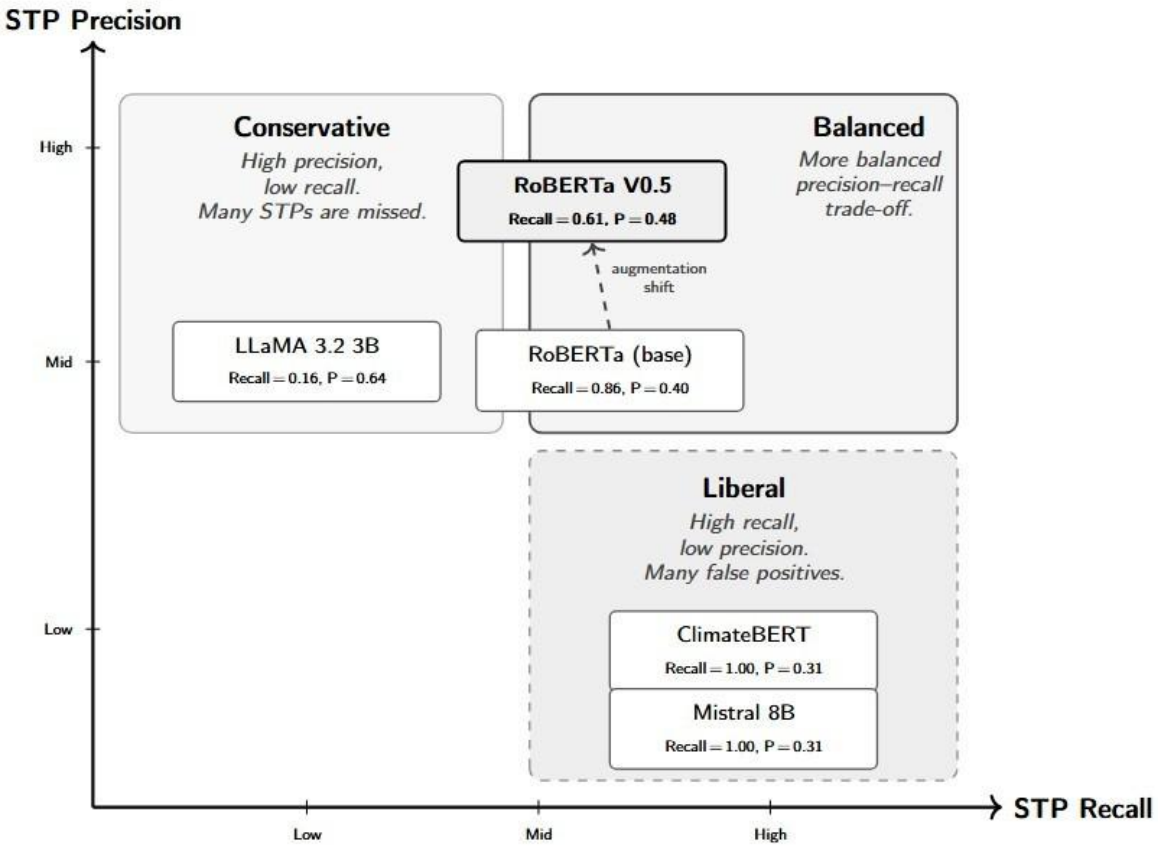

**Fig. 10** Where each model sits in the tipping-point precision-recall space (Set B, 161 passages). Models fall into three groups. Cautious models with high precision but low recall miss most tipping points (top-left). Loose models with high recall but low precision flag too many false positives (bottomright). The balanced group (top-right) is the useful one for large-scale evidence work. The dashed arrow traces RoBERTa across the four augmentation rounds (V0.2 to V0.5), moving from the upper edge of the balanced group into its middle

## 4.3 Checking against human experts

Eight research teams joined a check, each rating nine passages from the test set. Six teams finished, which gave 51 ratings. Human raters marked 35.3% of passages as tipping points. RoBERTa marked 27.5%, and GPT-4 marked 25.5%. Both automated tools were more cautious than the expert group.

The participating reviewers represented multiple disciplines including climate science, environmental policy, social systems analysis, and artificial intelligence. This diversity was intended to reflect the interdisciplinary nature of social tipping point research and to reduce bias associated with a single-domain assessment perspective.

Table 5 gives the result against the human agreement. RoBERTa reached 80.0% accuracy with tipping-point F1 of 0.69. GPT-4 reached 90.0% accuracy with tippingpoint F1 of 0.84 and perfect tipping-point precision, which means every passage GPT-4 called a tipping point was also called one by the human raters. The two automated tools agreed with each other 86.3% of the time. RoBERTa's confidence scores fell into two clusters, one near the decision line (0.5 to 0.6) and one high (above 0.85). This split tracks real difficulty. On five passages where RoBERTa was unsure (confidence 0.512 to 0.581), GPT-4 was sure on all five, yet human raters agreed with RoBERTa on three of them. RoBERTa's hesitation seems to follow real doubt in the text rather than a weakness in the model. The gap between GPT-4 and the open tuned encoder is real, but it is smaller than the gap in running cost between the two at scale, a trade-off taken up in Section 5.

**Table 5** RoBERTa and GPT-4 against the human expert agreement (n=51 passages). Support is the number of passages in each class by the agreed label

| System | Class | Precision | Recall | F1 | Support |
|---|---|---|---|---|---|
| RoBERTa | Non-tipping | 0.81 | 0.91 | 0.86 | 33 |
| | Tipping | 0.79 | 0.61 | 0.69 | 18 |
| | Accuracy | | | 0.80 | 51 |
| GPT-4 | Non-tipping | 0.87 | 1.00 | 0.93 | 33 |
| | Tipping | 1.00 | 0.72 | 0.84 | 18 |
| | Accuracy | | | 0.90 | 51 |

## 4.4 Rewriting and rating results

The rewriting step ran on 57 detected tipping-point passages. It gave a mean length ratio of 1.15 (SD = 0.31), adding about 126 characters per passage. A quality check flagged 6 passages (10.5%) for growing more than 1.5 times the source length. Figure 11 shows the review panel for one such passage.

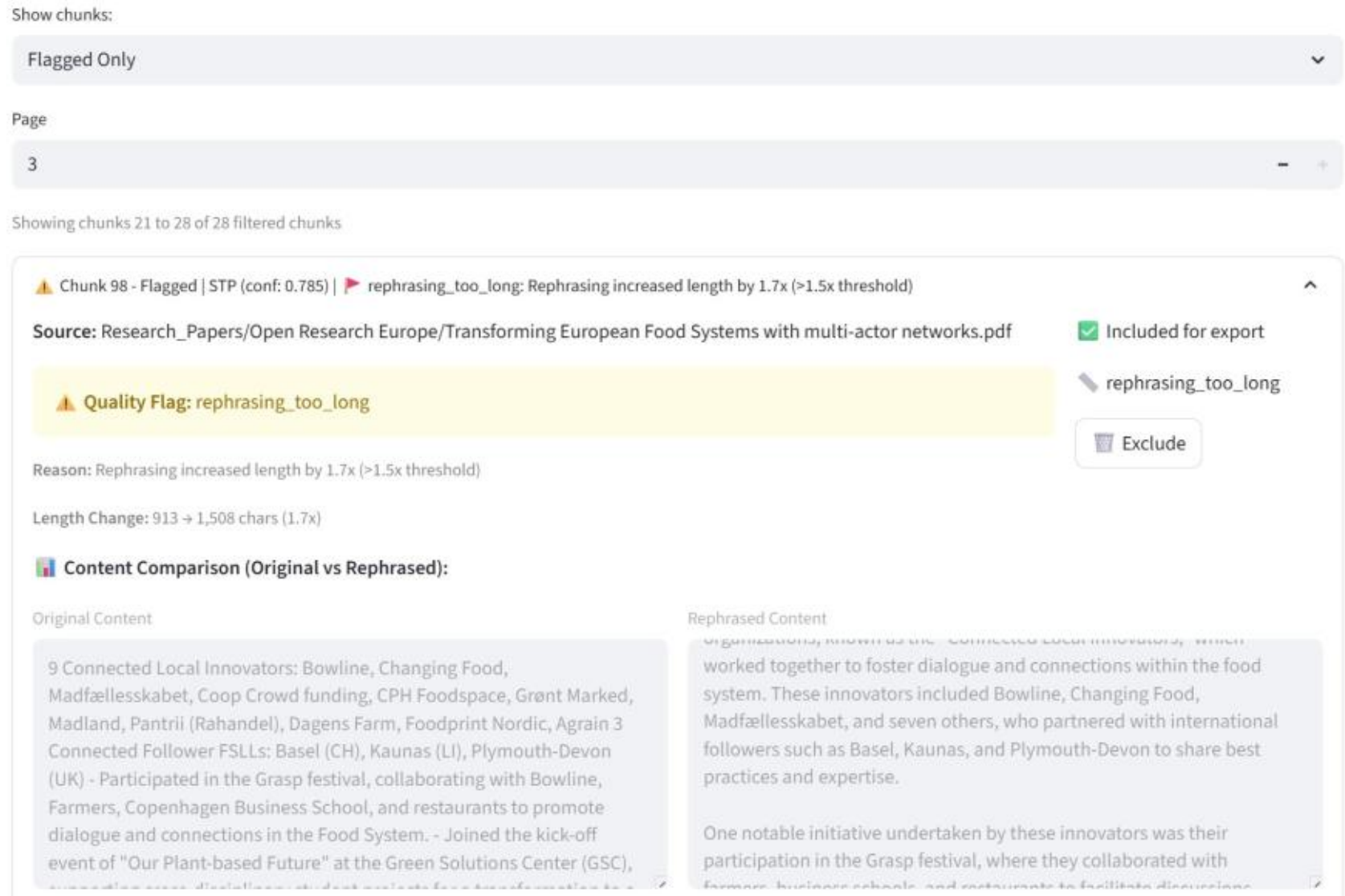

**Fig. 11** Quality review panel in the results tab, showing one flagged passage (Chunk 98, STP confidence 0.785). The rephrasing too long flag fires when a rewritten passage grows beyond 1.5 times the source length. The side-by-side view of the original and rephrased text lets a reviewer decide whether to include or exclude the passage before export. The right-hand panel shows the quality tag and the exclude button, which writes the decision back to the MinIO store so that downstream consumers see a consistent, reviewed result set

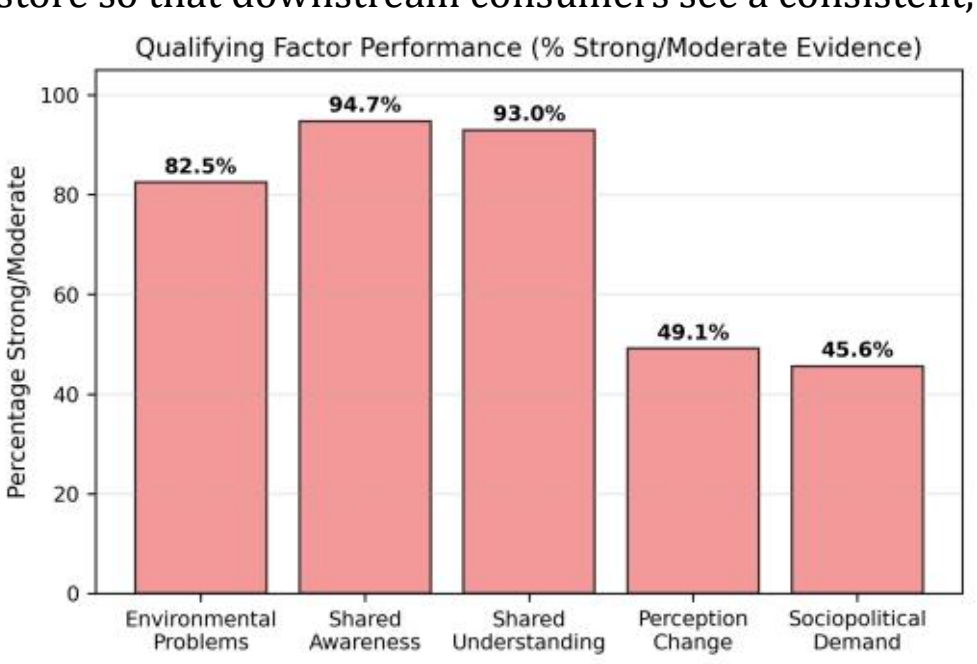

**Fig. 12** Rating results. Bars show the share of detected tipping-point passages (n=57) rated Strong or Moderate on each of the five criteria: felt social consequences, shared awareness, shared understanding, demand for habit change, and demand for political action. Shared awareness and shared understanding appear in over 90% of passages. Demand for habit change and demand for political action appear in fewer than half

Figure 12 shows the rating results. Shared awareness was rated Strong or Moderate in 94.7% of passages, and shared understanding in 93.0%. Authors of tipping-point work tend to describe what a community knows and understands before they present findings, so these two come up most often. Demand for habit change was lower at 49.1%, and demand for political action reached only 45.6%. This pattern reflects how these documents are written, since they more often set out the problem and its causes than push for a particular response.

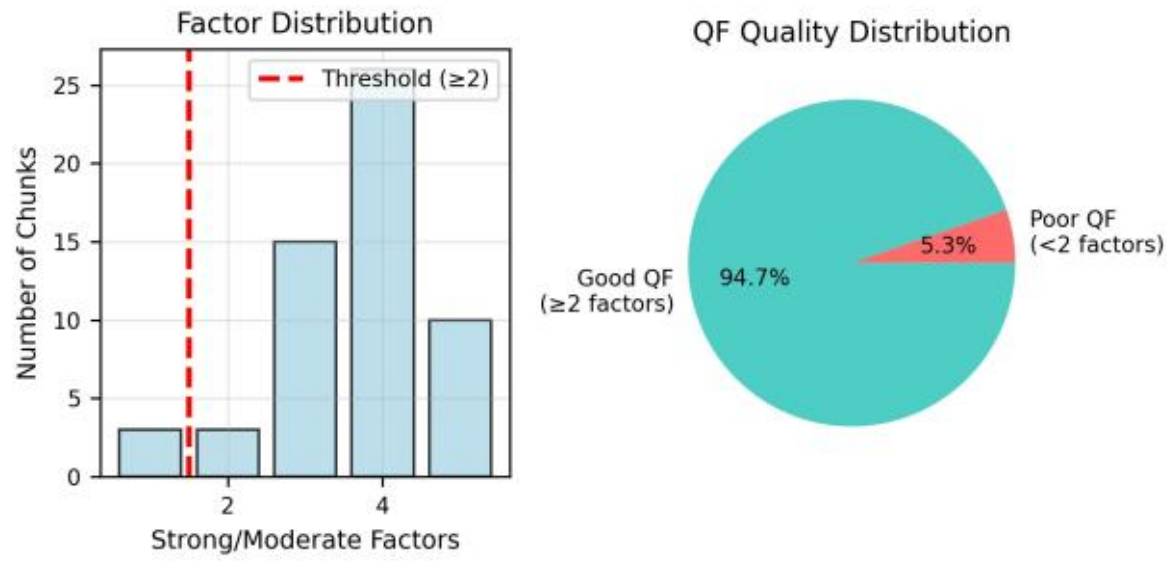

**Fig. 13** How the rating results spread across 57 tipping-point passages. Left: how many Strongor-Moderate criteria each passage had, with the quality bar at 2 criteria. Right: the overall quality check, showing that 94.7% of passages clear the bar for sound rating. The 5.3% below the bar are flagged as likely false positives from the sorter

Figure 13 shows that 94.7% of passages cleared the bar of at least two Strong-orModerate criteria. Of these, 45.6% had four such criteria and 26.3% had three. Only 5.3% fell below the bar, which flags them as likely false positives. The combined expert review queue, counting both the length flags and the below-bar ratings, covers 15.8% of detected passages.

Table 6 and Figure 14 sum up how closely the rewrite kept to the source. Mean cosine similarity was 0.84 (SD = 0.08), with 75% of rewrites above 0.80 and only three (5.3%) below 0.75. The moderate Jaccard similarity (mean 0.28) shows that the model truly reworded the content rather than copying it, changing both sentence shape and word choice while keeping the meaning. Lexical richness was the same in the source (mean 0.66) and the rewrite (mean 0.66), with a Pearson correlation of 0.78 ($p < 0.001$), which shows that the rewrite neither simplified nor padded the wording.

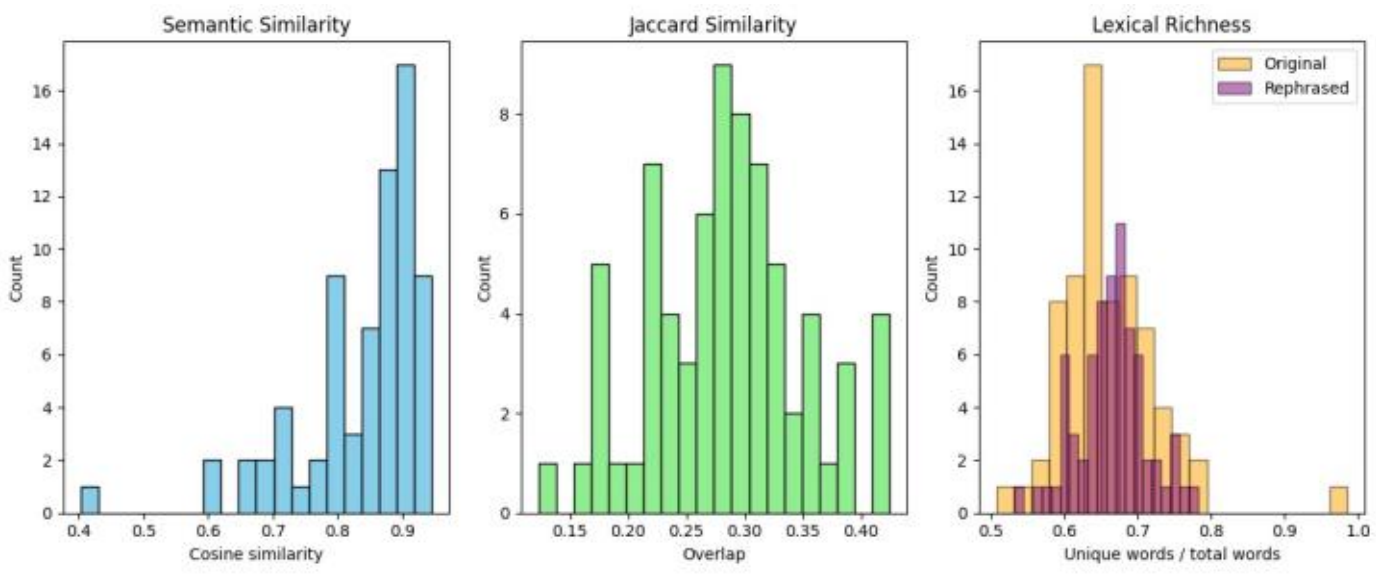

**Fig. 14** How closely rewriting kept to the source across 57 passages. Left: cosine similarity between source and rewritten passage, mean 0.84. Centre: Jaccard similarity, mean 0.28, the mark of a real rewrite rather than a near-copy. Right: lexical richness (type-token ratio), steady between source and rewritten text

**Table 6** How closely rewriting kept to the source (n=57 passages). High cosine similarity with moderate Jaccard similarity is the mark of a real rewrite, not a near-copy. Lexical richness held steady between source and rewritten text

| Measure | Mean | SD | Range |
|---|---|---|---|
| Cosine similarity | 0.84 | 0.08 | 0.65–0.96 |
| Jaccard similarity | 0.28 | 0.06 | 0.16–0.42 |
| Lexical richness (source) | 0.66 | 0.07 | 0.51–0.82 |
| Lexical richness (rewritten) | 0.66 | 0.06 | 0.53–0.78 |

# 5 Discussion and conclusion

## 5.1 General discussion

This study set out to detect and structure social tipping point evidence in climate documents, and to deliver that capability as an open and modular system that a small team can run on its own computational infrastructure. The framework joins four model components with a shared object store and a shared vector store, and it gives each stage its own configuration panel so that a user can swap in a new model or tune a single threshold without touching the rest of the system. The components were tested separately. DistilBERT boundary splitting gave passages with better combined coherence and coverage than the topic-model, semantic-similarity, and fixed-window methods, with a combined score of 6.137 against the best rival score of 5.609. A general RoBERTa with task tuning beat the climate-trained ClimateBERT and the untuned language models on balanced detection, reaching kappa 0.337 on the full benchmark and 0.742 on clear cases. The four augmentation rounds raised the score overall, though not at every step, and the third-round dip exposed a class-balance effect specific to very small training sets.

Section 2 placed this work against earlier climate text systems in Table 1. As can be seen, early systems sort whole documents by topic or flags a single factual claim in company text, and none reads passage-level concepts in long-form climate documents or builds social tipping features into its training target. The framework presented in this article fills that gap and works on the broader climate record, including the research papers that Callaghan et al. [4] and Berrang-Ford et al. [14] map at the document level. The tools fit together rather than compete. A full evidence workflow could use a Callaghan-style mapper to find the right documents, the framework presented here to pull and structure tipping-point evidence inside them, and a fact-checking system in the style of Leippold et al. [52] to test specific claims. The result is a reusable starting point for tipping-point evidence work at a scale beyond hand coding.

## 5.2 Why the splitting method matters

The splitting result backs the finding of Wang et al. [48] that splitting quality drives later task performance. The trade-off it exposes is between coherence and coverage. Tight similarity gives clean passages but leaves 39% of the document out, which means that 39% of tipping-point candidates would never reach the sorter. The larger passages from BERTopic mix tipping-point and ordinary content, which weakens the signal. The fixed window did worst, which confirms that these documents do not switch topic at set points, the pattern fixed windows were built for [44]. The DistilBERT method, with its mix of high coverage and good coherence, is the best practical choice for this kind of text and a sound default for climate evidence systems built on long-form documents. This fits the recent finding that splitting choice strongly shapes retrieval quality in climate question-answering [3].

For anyone building a system of this kind, the lesson is that the splitting choice should be a first-class design decision rather than an afterthought. It sets both how much of a document reaches the sorter and how cleanly the sorter's input holds one topic, which is why the splitter is exposed as its own configuration panel in the present system.

## 5.3 Why a general model beats a climate-specific one

The classification result extends the comparison of Trajanov et al. [34]. On the tipping-point task, ClimateBERT's climate training hurts rather than helps. The model produces a degenerate positive prediction (recall 1.00, precision 0.31). The most likely cause is that ClimateBERT, trained on millions of ordinary climate paragraphs, learned to tie climate wording to the positive label, and the tipping-point signal at 90 examples is too weak to undo that lean. The same shows up in untuned Mistral 8B, where the one-sided lean is the main error. A general model like RoBERTa starts with no such lean and reaches better balance. This result lines up with the broader argument of Yan [38] that standard two-stage pre-training-then-fine-tuning does not always close the gap between a general source domain and a narrow target domain, and that an extra adaptation stage may be needed when the target task differs in shape from the pre-training corpus.

The human-check gap between GPT-4 (90% against the expert agreement) and the improved RoBERTa (80%) fits the wider finding that GPT-4-scale models apply hard definitions more steadily [3, 52]. The real question is cost. Running a GPT-4 pipeline costs far more per passage than running an open tuned encoder. For collections the size of Callaghan et al. [4] or Berrang-Ford et al. [14], the tuned open model is the workable choice, and a GPT-4-scale model is best kept for a second-pass check [41, 42].

The wider point for practice is that climate-specific training is not always the better choice for a narrow concept task with little data. The choice should be tested, not assumed, and the evidence here favours a general model for tipping-point passage classification at this data size.

## 5.4 Uneven Gain Progression Across Augmentation Rounds

On Set B, Cohen's kappa increased from 0.284 to 0.337 across four augmentation rounds, with a transient decline to 0.270 at the third round. The temporal-confusion fix sharpened the line between slow and fast change but moved the decision line in a way that hurt class balance. Because there are only 90 positive examples, each synthetic batch is a large share of the training data, and a fix aimed at one error can bring in a new tendency. This cross-round effect was not foreseen by the first error analysis.

The result adds to the targeted-augmentation work of Jain et al. [40] and Wei and Zou [39]. On a very small training set, the order and make-up of the rounds matter, and watching a balanced measure such as kappa across rounds, rather than accuracy alone, is needed to catch a hidden shift in the boundary. The Set A picture gives a clearer version of the same point. V0.2 is the peak on Set A, and the later rounds, tuned for hard cases, give back ground on clear ones. Each round moves the line to help one group at the cost of another, and the rounds together give a better but uneven result. The pattern also fits the wider point in Yang [43] that surface statistical patterns alone do not give a stable signal when the training set is small, so a fix that shifts the data distribution can move the decision boundary in ways that are hard to anticipate.

In practical terms, a larger expert-labelled set would lower the sway of any one batch and likely give consistent gains per round. In the small-data setting, though, tracking kappa across rounds is itself a useful check for shifts that accuracy alone hides. The system exposes both V0.2 and V0.5 ONNX classifiers through the configuration interface so that a user can pick the right model for the task at hand.

## 5.5 Implications for applied pipeline design

The findings carry implications for the design of applied text mining systems for evidence synthesis at scale, beyond the social tipping point case taken up here.

The first is that a modular design pays off in terms of cost and reuse. The framework is built from four model components that can be swapped on their own, and each was tested against alternatives so that the choice of part is grounded rather than assumed. As benchmarks such as ClimateEval [18] grow and better domain models appear, the classification step can be replaced by changing the path to the new ONNX file without touching the splitter or the rewriter, and the rating frame stays separate from whichever model runs it. The same modular logic supports the open data-catalogue approach of Ouaknine et al. [36] for forest monitoring, and it lets later teams reuse parts rather than rebuild end-to-end systems.

The second is that the deployment cost gap between an open tuned encoder and a closed large language model is the gating factor for scale. The human-check gap between GPT-4 (90% accuracy) and the tuned RoBERTa (80%) is real, but the cost gap per passage at the scale of Callaghan et al. [4] or Berrang-Ford et al. [14] is much larger. The open tuned encoder is the workable choice for first-pass detection across a large corpus, and a GPT-4-scale model is best kept for a second-pass check on the harder cases [41, 42]. Recent reviews of model compression [51] show that running tuned encoders on resource-constrained hardware is a tractable engineering problem rather than a fixed barrier, which makes the open-encoder route practical for teams that lack a sufficient resources.

The third is that the rating step puts a person back in the loop where the judgement is hardest. The rating routes 15.8% of detected passages to a human review queue, and the classifier's own uncertainty tracks genuine difficulty in the text rather than noise, as the human check showed. A working team can read the hard cases and trust the clear ones. This matters most for a concept like social tipping, where the line between a tipping point that people can act on and ordinary fast change is still contested in the field [13, 19]. A tool that hid its uncertainty would risk hardening a contested definition into whatever the model happened to learn. The modular build protects against this in a second way, since the rating frame can be updated as the definition tightens, without having to retrain any of the upstream components.

The fourth limitation concerns the completeness of the evidence record. The system captures what is explicitly stated in the processed documents rather than what is absent or understated, and the distribution of qualifying factor ratings reflects this constraint. Shared awareness and shared understanding appear in over 90% of detected passages, while demand for action appears in fewer than half. The model is not missing this evidence. The documents in this corpus more often set out a problem and its causes than push for a specific response. A responsible use of the system reads its outputs against this gap rather than treating absence as confirmation that no demand exists.

The fifth is that wrapping the model components in a Streamlit interface backed by a shared object store and a shared vector store turns a research pipeline into a tool a small team can run end to end. The configuration screenshot in Figure 7 shows that the user-facing surface of the system is a small set of named panels, each tied to one stage of the pipeline. The same pattern should generalise to other passagelevel concept detection tasks in environmental, health, and policy domains, where the target concept is rare in any one document but spread across a large corpus.

## 5.6 Limitations and future research

The training set of 90 positive examples is small and covers only a few domains (floods, air pollution, over-tourism, plastic pollution). The model may not carry over well to tipping-point cases in domains not in the training set, such as urban policy, financial markets, or public health.

In addition, a substantial portion of the expanded training data was augmented through model data generation. Although this approach improved class balance and model performance, synthetic examples may not fully capture the diversity and complexity of real-world social tipping point narratives. The test labels were made by GPT-4.1, not by human domain experts, and the agreement between GPT-4.1 and human labels was not measured. The test numbers should be read as performance against a benchmark labelled by a language model, with the 51-passage human check as a partial cross-check.

The performance gap between the high-confidence and mixed-confidence evaluation sets suggests that classification remains challenging for borderline cases where even the benchmark labels carry greater uncertainty.This observation is consistent with the lower Cohen values K obtained in the mixed-confidence data set, indicating reduced agreement on ambiguous passages. The human check covered only 51 passages from a limited set of source documents, which restricts statistical power and may not fully represent the diversity of climate and environmental narratives encountered in practice.

The rewriting and rating steps were not checked against expert-written summaries. The fidelity measures show that the rewrite keeps the source meaning, but they do not show whether the ratings match expert judgement on hard passages.

The text enhancement evaluation focused on semantic preservation and lexical characteristics. While these measures indicate that content fidelity was maintained during rephrasing, readability and usefulness for end-users were not assessed through dedicated human evaluation and remain an important direction for future work.

Three steps would strengthen the system and shape the next stage of this work. A structured labelling campaign would grow the expert set and lower the reliance on synthetic data. A measure of agreement between GPT-4.1 labels and human labels would put the benchmark on firmer ground.

## 5.7 Conclusion

This paper presented a modular framework for detecting and structuring social tipping point evidence in climate-related documents at the passage level. The framework combines a DistilBERT boundary classifier for segmentation, a RoBERTa model tuned through targeted augmentation for classification, a Mistral 7B model for rewriting detected passages, a LLaMA 3.2 3B model for rating passages against five published tipping point criteria, and a Milvus vector store for retrieval, deployed through a Streamlit interface backed by MinIO storage.

The boundary classifier achieved the highest composite score among four segmentation methods on a nine-metric evaluation (6.137). It preserved tipping-point passage boundaries across varied document structures where fixed-window and topic-clustering approaches produced erroneous splits. The tuned RoBERTa classifier reached 71.4 percent accuracy with a Cohen's kappa of 0.337 on the full 163-passage benchmark, and 87.5 percent accuracy with a kappa of 0.742 on the subset of clearly labelled passages, outperforming both a climate-specific model (ClimateBERT) and a larger untuned language model (LLaMA 3.2 3B). Four rounds of targeted augmentation, each addressing a specific classification error identified through error analysis, improved class balance while keeping the ratio of real to synthetic training examples fixed at 70:30. The rewriting and rating stages converted detected passages into structured evidence records suitable for expert or policy review.

The framework has several limitations. The training set of 90 positive examples covers a limited set of domains, and a substantial share of the training data is synthetic rather than expert-labelled. Test labels were produced by GPT-4.1 rather than verified against a full set of human annotations, and the 51-passage human check covers only a subset of the test data. The performance gap between the full benchmark and the highconfidence subset indicates that classification accuracy depends on label certainty, and this should be taken into account when interpreting the reported results.

# Statements and Declarations

- **Funding.** Funding information is provided on the separate title page and entered in the submission system to preserve double-anonymous review.

- **Competing interests.** The authors have no relevant financial or non-financial interests to disclose.
- **Ethics approval and consent to participate.** Not applicable. The study analyses text passages from published scientific literature and expert annotations of those passages. It does not involve human participants, their personal data, or identifiable information.
- **Consent for publication.** Not applicable.
- **Data availability.** The STP dataset collected in this study is currently in the process of being formally published and will be made available upon completion of that process.
- **Materials availability.** Not applicable.
- **Code availability.** The code used in this study is not publicly available.
- **Use of large language models.** Large language models are part of the study's method and are documented in Section 3: GPT-4.1 produced the benchmark labels, GPT-4 and Claude Sonnet produced the targeted augmentation batches, a Mistral 7B model performed the rewriting, and a LLaMA 3.2 3B model performed the criterion rating. Large language models were not used to design the study or to draft its scientific content. Any use of an AI tool in preparing the manuscript was limited to grammar and language checking, and the authors take full responsibility for the final text.